\documentclass{article}
\PassOptionsToPackage{numbers,compress}{natbib}

 \usepackage[eandd, final]{neurips_2026}

\usepackage[utf8]{inputenc} 
\usepackage[T1]{fontenc}    
\usepackage{url}            
\usepackage{booktabs}       
\usepackage{amsfonts}       
\usepackage{nicefrac}       
\usepackage{microtype}      
\usepackage{xcolor}         
\usepackage{pifont}
\usepackage{subcaption}
\usepackage{amsmath}

\usepackage{arydshln}
\usepackage{graphicx}

\usepackage{makecell}
\usepackage{multirow}
\usepackage{colortbl}
\usepackage{array}
\usepackage{wrapfig}
\usepackage{amssymb}

\newcommand{\tabincell}[2]{\begin{tabular}{@{}#1@{}}#2\end{tabular}}

\definecolor{neuripsblue}{rgb}{0.12,0.49,0.85}
\definecolor{lavender}{RGB}{180,140,200}
\definecolor{myPurple}{RGB}{251,237,255}
\usepackage[pagebackref=false,breaklinks,colorlinks,citecolor=neuripsblue]{hyperref}

\def\eg{{\em e.g.}}
\def\ie{{\em i.e.}}

\definecolor{mygray}{gray}{0.9}
\definecolor{highlight}{RGB}{238,250,215}
\definecolor{lightbluegray}{HTML}{8CBBED}

\title{Towards Visual Query Segmentation in the Wild}

\author{%
  Bing Fan$^*$ \; Minghao Li$^*$ \; Hanzhi Zhang$^*$ \; Shaohua Dong \; Naga Prudhvi Mareedu \\ \textbf{Xiaoqiong Liu} \;\; \textbf{Weishi Shi} \;\; \textbf{Yunhe Feng} \;\; \textbf{Yan Huang} \;\; \textbf{Heng Fan}$^\dag$ \\
  Department of Computer Science and Engineering, University of North Texas\\
  $^*$Equal contribution \;\;\; $^\dag$Corresponding author\\
}

\begin{document}

\maketitle

\begin{abstract}
    In this paper, we introduce \emph{visual query segmentation} (VQS), a new paradigm of visual query localization (VQL) that aims to segment \emph{all pixel-level} occurrences of an object of interest in an \emph{untrimmed} video, given a visual query. Compared with existing VQL locating only the last appearance of a target using bounding boxes, VQS enables more comprehensive (\ie, \emph{all} object occurrences) and precise (\ie, \emph{pixel-level} masks) localization, making it more practical for real-world scenarios. To foster research on this task, we introduce VQS-4K, a large-scale benchmark dedicated to VQS. Specifically, VQS-4K consists of 4,111 videos with more than 1.3 million frames, and covers a diverse set of 222 object categories. Each video is paired with a visual query defined by a frame \emph{outside} search video and its target mask, and annotated with spatial-temporal masklets corresponding to the queried target. To ensure high quality, all videos in VQS-4K are manually labeled with  meticulous inspection and iterative refinement. To our best knowledge, VQS-4K is the \emph{first} benchmark specifically designed for VQS. Furthermore, to stimulate future research, we present a simple yet effective method, named VQ-SAM, which extends SAM 2 by leveraging target-specific and background distractor cues from the video to progressively evolve the memory through a novel multi-stage framework with an adaptive memory generation (AMG) module for VQS, significantly improving the performance. In our experiments on VQS-4K, VQ-SAM achieves promising results and surpasses existing methods, evidencing its effectiveness. With VQS-4K and VQ-SAM, we expect to inspire more future research and applications on VQS. Our benchmark and code are released at \url{https://github.com/VAIL-Fan/VQS}.
  
\end{abstract}

\section{Introduction}
\label{sec:intro}

Visual query localization (VQL) aims at locating an object of interest, in \emph{space} and \emph{time}, within an \emph{untrimmed} video, given a visual query obtained outside the search video. As a fundamental computer vision task, it has many applications such as surveillance, robotics, and video retrieval and editing.

\begin{figure}[!t]
    \centering
    \includegraphics[width=0.99\linewidth]{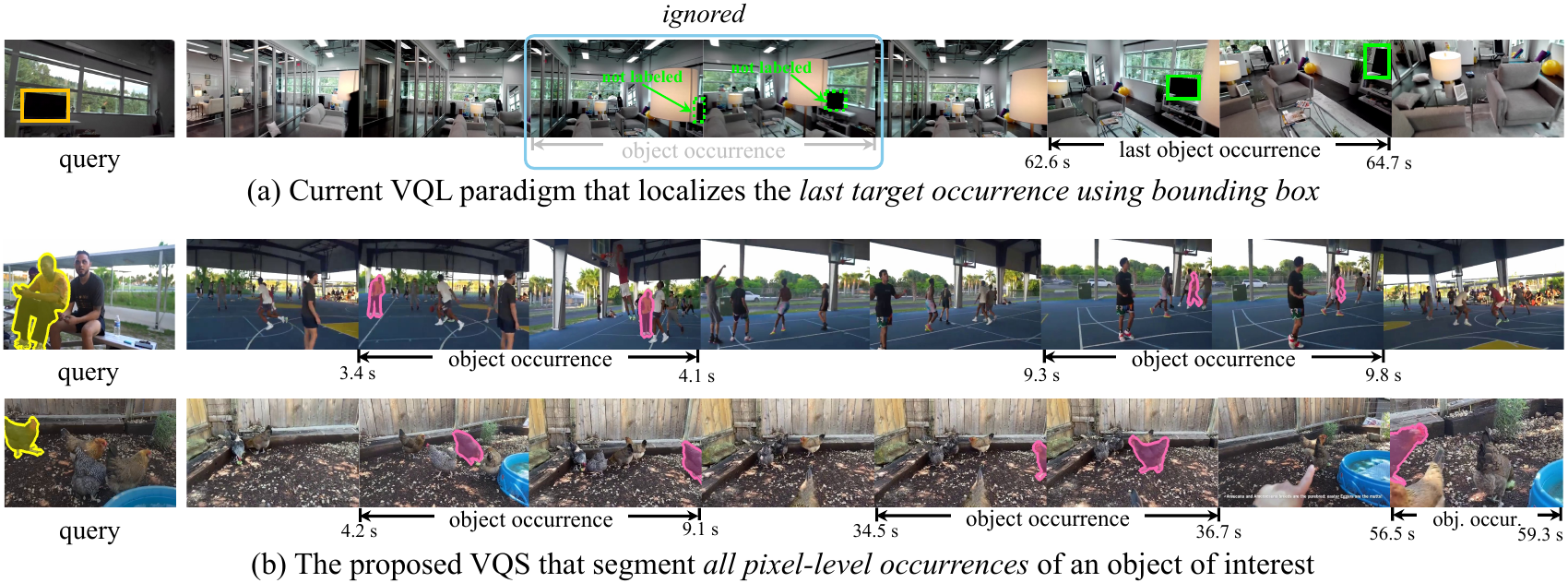}
    \caption{Comparison of existing VQL paradigm in image (a) and the proposed VQS in image (b).}
    \label{fig:task}
\end{figure}

Currently, the research of VQL mainly focuses on locating the \emph{last} target appearance in a video~\cite{grauman2022ego4d} (see Fig.~\ref{fig:task} (a)). Despite progress~\cite{jiang2023single,xu2023my,khosla2025relocate,chang2025hero,fan2025prvql,manigrasso2026online,khosla2025ren}, this VQL paradigm remains \emph{insufficient} for many real-world scenarios involving multiple target occurrences, such as visual surveillance and video object retrieval, in which knowing \emph{all} target occurrences, rather than only the last one, is essential for comprehensive video understanding. In addition, in the current VQL setting, objects are often represented using \emph{bounding boxes} (see again Fig.~\ref{fig:task} (a)). While simple, this representation tends to introduce noise into objects, and thus limits precise spatial localization of targets, hindering its applicability to downstream tasks demanding accurate target localization such as video editing.

To mitigate the above issues, we present \emph{visual query segmentation} (VQS), a new VQL paradigm that aims to segment \emph{all pixel-level} occurrences of a target of interest within an untrimmed video, given a visual query (see Fig.~\ref{fig:task} (b)). Compared with existing VQL~\cite{grauman2022ego4d}, VQS enables both \emph{comprehensive} localization by capturing \emph{all} target occurrences in a video and \emph{precise} localization by representing objects with \emph{pixel-level} masks, making it more suitable in real-world applications. It is worth noting that, despite sharing certain similarities to video object segmentation (VOS)~\cite{yang2019video,perazzi2016benchmark,xu2018youtube,ding2023mose,hong2023lvos}, VQS fundamentally differs. Unlike VOS where the reference target is sampled from the first frame \emph{within} the video, the query in VQS is sourced from an image frame \emph{outside} the search video. This makes localization in VQS more challenging, as there may be no exact visual match for the queried object. Besides, VOS typically segments the target via sequential propagation in a \emph{trimmed} video, whereas VQS aims to localize and segment the queried object across an \emph{untrimmed} video, requiring the model to identify sparse target occurrences amidst substantial background distractors.

Considering the lack of a dedicated benchmark for studying VQS, we propose VQS-4K, a large-scale dataset for this task. VQS-4K comprises 4,111 videos with over 1.3 million frames, spanning 222 categories across diverse in-the-wild contexts. Each video is paired with a visual query that is sampled from a frame outside the search video, together with a mask indicating the target of interest, and is annotated with a varying number of spatio-temporal masklets that correspond to the queried target. To ensure the high quality of our VQS-4K, all mask annotations in each sequence are manually annotated with meticulous inspection and undergo multi-round careful refinement as necessary. To our best knowledge, VQS-4K is to date the \emph{first} benchmark specifically designed for visual query segmentation. Fig.~\ref{fig:task} (b) demonstrates several example sequences in VQS-4K. By releasing VQS-4K, we aim to offer a dedicated platform to the community for facilitating the exploration of VQS. 

Furthermore, to encourage future research, we present a simple yet effective model for VQS, termed VQ-SAM, by extending SAM 2~\cite{RaviGHHR0KRRGMP25}. Its core is to leverage both target-specific information and background distractor cues to progressively evolve the memory for VQS. Concretely, VQ-SAM is formulated as a multi-stage architecture. In each stage, we first generate potential object masks from a video using current memory. With these masks and confidence scores, we then select a few top target and distractor regions to extract the target-specific and distractor features respectively, similar to~\cite{videnovic2025distractor}. These features, together with initial memory from the visual query, are utilized to generate the new memory. Considering that different features may contribute unequally in memory, we propose an adaptive memory generation (AMG) module that learns importance weights for different feature types and integrates them accordingly to form memory. By incorporating the target-specific features, the newly evolved memory can better handle target variations in videos. Besides, leveraging distractor features enables more reliable discrimination between the target and the background. In VQ-SAM, the memory generated in one stage is used in the next stage to produce more accurate target-specific and background distractor features, which are then used to further evolve the memory. Through this progressive process, the memory is gradually improved in VQ-SAM, resulting in better localization and segmentation in the final stage. To validate VQ-SAM, we conduct extensive experiments on VQS-4K. The results show that, our VQ-SAM demonstrates promising performance and largely surpasses current methods, establishing a solid foundation for future research on VQS.



In summary, we make the following contributions: \ding{171} We propose visual query segmentation (VQS), a new VQL paradigm which segments \emph{all pixel-level} target occurrences; \ding{170} We introduce VQS-4K, a large-scale dataset containing 4,111 videos for VQS; \ding{168} We present VQ-SAM, a simple yet effective approach to stimulate future research of VQS; \ding{169} We show VQ-SAM achieves promising results, expecting it to serve as a reference and provide guidance for future algorithm design.

\section{Related Work}
\label{related_work}


\textbf{VQL Benchmark.} Benchmark is crucial for training and evaluating VQL algorithms. To date, VQ2D, introduced in Ego4D~\cite{grauman2022ego4d}, is the only available dataset tailored to this task. It contains 2,178 long first-person videos, which are divided into 5,842 clips under the current VQL paradigm, mainly focusing on locating the final target appearance using bounding boxes. \emph{\textbf{Different from}} VQ2D, VQS-4K is specially developed for the new VQS task for more comprehensive and precise target localization. To this end, it provides pixel-level mask annotations for all target occurrences throughout the video (see Fig.~\ref{fig:task} (b) again). Moreover, \emph{\textbf{unlike}} VQ2D, which mainly focuses on locating rigid target objects from an egocentric viewpoint, VQS-4K aims to localize arbitrary targets, rigid or deformable, across both first- and third-person views, which makes it more challenging yet practical in real-world scenarios.

\textbf{VQL Algorithm.} VQL algorithms have recently witnessed rapid progress. The work of~\cite{xu2023my} proposes a two-stage framework that first detects the target and then performs bi-directional tracking from the detection result for spatial-temporal localization. To reduce compounding errors across different stages and to improve efficiency, the work of~\cite{jiang2023single} introduces a single-stage Transformer architecture for VQL, showing promising performance. Building on~\cite{jiang2023single}, the method of~\cite{fan2025prvql} exploits target cues in the video to refine features for more accurate localization, while the model in~\cite{chang2025hero} explores attention guidance and egocentric augmentation to enhance performance. In~\cite{khosla2025relocate}, a training-free framework is presented to exploit foundational models for VQL, and shows excellent results. This approach is later extended in~\cite{khosla2025ren} with a better feature network. Unlike these methods, the work of~\cite{manigrasso2026online} introduces an online VQL approach. \textbf{\emph{Different from}} the above methods localizing the last target appearance with boxes as in existing VQL paradigm, VQ-SAM is specially designed for the new VQS task to segment all target occurrences at pixel-level, enabling more comprehensive and precise target localization.


\textbf{Video Object Segmentation (VOS).} VOS is a crucial vision task, aiming to separate foreground and background pixels in a video given the reference target, and has been extensively investigated in the past decade with many datasets (\eg,~\cite{yang2019video,perazzi2016benchmark,xu2018youtube,ding2023mose,hong2023lvos,qi2022occluded}) and methods (\eg,~\cite{oh2019video,cheng2022xmem,wu2023scalable,cheng2024putting,yan2024visa,qin2025structure,liu2025livos,RaviGHHR0KRRGMP25}) proposed. Similar to VOS, VQS segments the foreground target from background in a video. However, a key \textbf{\emph{difference}} between them lies in the source of reference target. In VOS, the reference is directly obtained from the first frame \emph{within} the search video, whereas in VQS, the reference target, or the visual query, is sampled from a frame \emph{outside} the search video, making VQS more challenging. Besides, \emph{\textbf{unlike}} VOS, which typically performs sequential \emph{frame-level} propagation in a \emph{trimmed} video, VQS requires \emph{video-level} search from an \emph{untrimmed} video, which increases the task complexity and necessitates different methodological designs.

\section{The Proposed VQS-4K}

\subsection{Design Principle}

We aim to provide a dedicated platform for visual query segmentation. In constructing the dataset, we adhere to the following key principles: (1) \emph{\textbf{Dedicated benchmark.}} One of the key motivations of this work is to provide a dedicate dataset for visual query segmentation. To align with the goal of VQS, each video sequence is paired with an external visual query, and annotated with spatial-temporal masklets corresponding to the query. Besides, to enable large-scale training and evaluation of models in current deep learning era, we expect the dataset to contain at least 4,000 videos. (2) \emph{\textbf{Diverse object categories.}} Diversity is crucial for a VQS dataset to support the development of generalizable VQS systems that are able to localize the target of arbitrary classes. To construct a diverse platform for VQS, we expect the new dataset to include at least 200 categories across diverse in-the-wild contexts. (3) \emph{\textbf{High-quality annotation.}} High-quality annotation is crucial for the dataset in both model training and evaluation. In order to ensure the high quality of our benchmark, each sequence is manually labeled with precise masklets for all target occurrences through iterative inspection and refinement. 

\subsection{Data Acquisition}

Our dataset aims to facilitate the development of generalizable VQS by covering rich classes drawn from a wide range of real-world contexts. To this end, a diverse set of 222 object categories are selected in our VQS-4K. Following many existing video datasets (\eg,~\cite{yao2025omnistvg,peng2024vasttrack}), the classes in VQS-4K are sourced from the popular ImageNet~\cite{deng2009imagenet} and V3Det~\cite{wang2023v3det} datasets, and organized in a hierarchical structure. It is worth noting that, all selected categories in our VQS-4K are verified by the domain experts (\eg, PhD students who work on related topics) to ensure their appropriateness for VQS. Compared to VQ2D, the classes in VQS-4K are more diverse and more desirable for visual query localization. Due to space limitation, we show detailed classes in VQS-4K in the \emph{appendix}.

After determining the categories in VQS-4K, we then start searching for video sequences of each category from YouTube, the largest and the most popular video platform with massive real-world videos from various scenarios. Throughout the search process, only videos under the creative commons license are collected, and they are used exclusively for the research purpose. Initially, we collect more than 8K videos using keywords relevant to object classes. Then, we carefully inspect each video to verify its availability for VQS. Specifically, if a suitable pair that consists of a video clip and an external query outside the clip exists for our task, we keep it; otherwise, we discard it. After that, for each qualified video sequence, we select the identified pair of clip and query for our benchmark, and gather a total of 4,111 video-query pairs for VQS. 

Eventually, we compile a large-scale dataset, termed VQS-4K, for VQS. VQS-4K contains 4,111 videos with 1.3 million frames and encompasses 222 categories from broad contexts. Each video is paired with an external visual query, and contains one or multiple target responses corresponding to the query. The average video length and the object response length are 52.9 and 9.4 seconds, respectively, at a frame rate of 6 frames per second (fps). Compared to VQ2D~\cite{grauman2022ego4d}, which is the only benchmark for existing VQL paradigm, our VQS-4K offers more comprehensive and precise mask information as well as more diverse object categories for VQS. Tab.~\ref{tab:comp} summarizes our VQS-4K and its comparison to VQ2D~\cite{grauman2022ego4d} and representative related VOS datasets~\cite{xu2018youtube,ding2023mose,hong2023lvos}.

\renewcommand{\arraystretch}{1.1}
\begin{table}[!t]
\setlength{\tabcolsep}{3pt}
  \centering
  \caption{Summary of VQS-4K and its comparison with VQ2D and representative VOS datasets. ``n/a'' denotes that the statistic is not applicable to the dataset. LO: Last-Occurrence; AO: All-Occurrence; IV: Inside Video; OV: Outside Video. Please note that, the frame rates of all datasets are 6 \emph{fps}.}
  \resizebox{0.99\textwidth}{!}{
    \begin{tabular}{rccccccccccc}
    \Xhline{1.2pt}
    \rowcolor[HTML]{f2f3f4} Benchmark & 
    \tabincell{c}{Task}&
    \tabincell{c}{Videos} & \tabincell{c}{Object \\ classes} & \tabincell{c}{Avg. vid. \\length} & \tabincell{c}{Avg. res. \\length} & \tabincell{c}{Total \\ frames} &  \tabincell{c}{Anno. \\ frames} & \tabincell{c}{Mean \\ occur.} & \tabincell{c}{Loc. \\ aim} & \tabincell{c}{Refer. \\ target} & \tabincell{c}{Object \\ anno.} \\
    \hline\hline
    \textbf{YT-VOS}~\cite{xu2018youtube}  & VOS & 3,252      &   94    &   3.6 sec    &   n/a    & 0.11M & 110K &    n/a & n/a   & IV   & mask  \\
    \textbf{LVOS}~\cite{hong2023lvos}  & VOS &  220     &    27   &   95.4 sec    &  n/a     &  0.13M & 130K &  n/a & n/a   & IV   & mask  \\
    \textbf{MOSE}~\cite{ding2023mose}  & VOS &  2,149     &  36     &   12.3 sec    &   n/a    & 0.16M & 160K &  n/a & n/a   & IV   & mask  \\
    \textbf{VQ2D}~\cite{grauman2022ego4d} & VQL &   5,823    &  135     &  366.0 sec     &    2.7 sec   & 12.0M  & 266K &    1 & LO  & OV  & box   \\
    \hline
    \textbf{VQS-4K} (ours) & VQS &   4,111    &  222  & 52.9 sec   &   9.0 sec  &  1.3M & 222K &  2.9  & AO  & OV & mask  \\
    \Xhline{1.2pt}
    \end{tabular}}
  \label{tab:comp}
  \vspace{-0.5em}
\end{table}%

\subsection{Data Annotation}

In VQS-4K, each video is labeled with spatial-temporal masklets that correspond to the given external visual query, enabling precise localization and segmentation of the queried object across time. Specifically, for each pair of video and query, we first identify all temporal segments in the video in which the queried object appears. These segments may be non-contiguous and span multiple occurrences of the target throughout the video. Subsequently, within each identified temporal segment, we perform dense annotations on the target of interest indicated by the visual query at the frame level using temporally consistent spatial masks.

To ensure the high quality of annotations, we employ an iterative labeling mechanism. Specifically, the experts first manually identify the start and the end timestamps of temporal segments for the queried target in each sequence. Then, the temporal segments of each video are manually labeled with masks by an annotation team formed by an expert and a few volunteers. After this initial round, the spatial-temporal mask annotations will be sent to a validation team of three experts for quality verification. If the annotations are not unanimously agreed by all experts, they will be returned back to the same labeling team for refinement. We repeat this process until annotations of all videos are completed. Due to space limiation, we show the pipeline of data annotation in the \emph{appendix}. Fig.~\ref{fig:task} (b) displays several annotation examples in VQS-4K, and more can be found in the \emph{appendix}.

\begin{figure}[!t]
    \centering
    \includegraphics[width=0.99\linewidth]{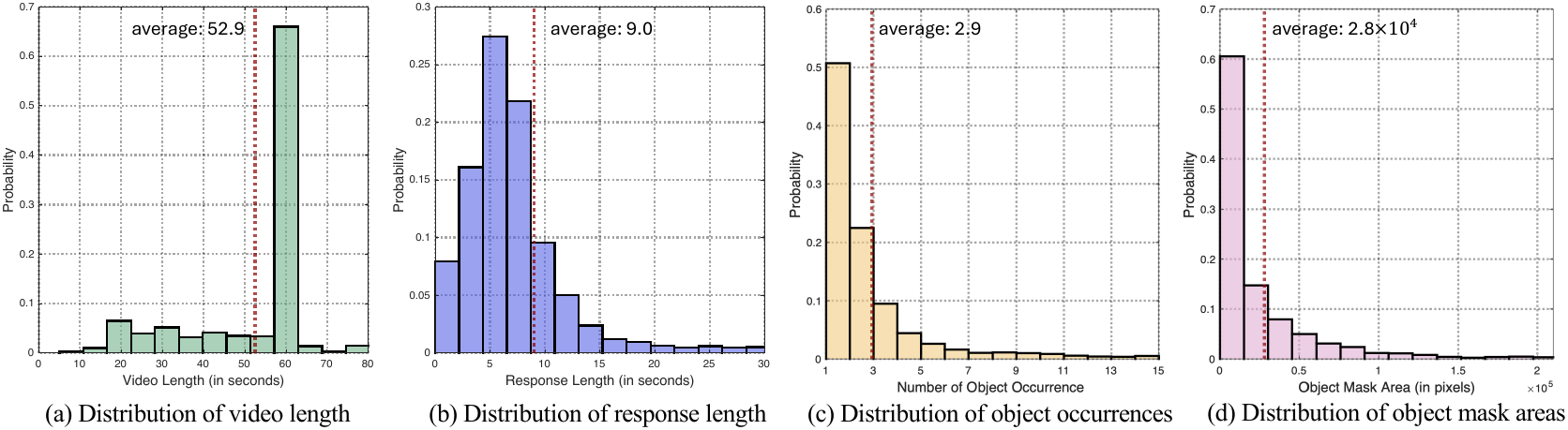}
    \caption{Representative statistics on VQS-4K. Please note, values greater than 80 in video length are grouped into the final bin for visualization purposes. }
    \vspace{-0.5em}
    \label{fig:stats}
\end{figure}

\textbf{Statistics.} To better understand our VQS-4K, we show representative statistics in Fig.~\ref{fig:stats}, including the distributions of video length, response length, number of object occurrences, and object mask areas. From Fig.~\ref{fig:stats}, we can observe that VQS-4K includes both long- and short-term videos, and individual videos often comprises multiple occurrences of the queried target. Together, these characteristics make the proposed VQS-4K well suited for real-world applications.

\subsection{Dataset Split and Evaluation Metric}

\setlength{\intextsep}{-1pt}
\setlength{\columnsep}{5pt}%
\begin{wraptable}{r}{0.5\textwidth}
\setlength{\tabcolsep}{1.5pt}
	\centering
	\renewcommand{\arraystretch}{1.1}
    \caption{Comparison of training and testing set.}
	\scalebox{0.82}{
    \begin{tabular}{lccccc}
    \Xhline{1.2pt}
    \rowcolor[HTML]{f2f3f4}
    & Videos & \tabincell{c}{Avg. Vid.\\ length} & \tabincell{c}{Avg. res. \\length} & \tabincell{c}{Total \\frames} & \tabincell{c}{Mean \\occur.} \\
    \hline\hline
    \textbf{VQS-4K$_{\text{Tst}}$} &   822    &   53.3 sec   &      6.9 sec  &  264K    &   2.7   \\
    \textbf{VQS-4K$_{\text{Tra}}$} &  3,289     &  52.8 sec    &   9.5 sec   &   1.04M    & 3.0  \\
    \Xhline{1.2pt}
    \end{tabular}}
	\label{tab:ttcomp}
    \vspace{12pt}
\end{wraptable}
\textbf{Dataset Split.} VQS-4K contains a total of 4,111 videos. Among them, 3,289 videos with 1.04M frames are used for the training set (dubbed VQS-4K$_{\text{Tra}}$), and the rest 822 sequences with 264K frames for the testing set (dubbed VQS-4K$_{\text{Tst}}$). In dataset split, both VQS-4K$_{\text{Tra}}$ and VQS-4K$_{\text{Tst}}$ contain the full 222 classes. Tab.~\ref{tab:ttcomp} shows the comparison of VQS-4K$_{\text{Tra}}$ and VQS-4K$_{\text{Tst}}$, and the detailed splits will be released with VQS-4K. 

Besides, to enable further analysis, we divide VQS-4K$_{\text{Tst}}$ into three subsets based on the average object mask area in each video, including VQS-4K$_{\text{Tst}}$-Small for \emph{small-scale} objects with average mask areas in the range of [0, 3.6$\times10^{3}$) pixels (269 videos), VQS-4K$_{\text{Tst}}$-Medium for \emph{medium-scale} objects with average mask areas in [3.6$\times10^{3}$, 4.0$\times10^{4}$) pixels (393 videos), and VQS-4K$_{\text{Tst}}$-Large for \emph{large-scale} objects with average mask areas greater than 4.0$\times10^{4}$ pixels (164 videos).

\vspace{0.3em}
\noindent
\textbf{Evaluation Metric.} We employ multiple metrics on VQS-4K for evaluation, comprising \emph{mean spatial-temporal IoU} (mSTIoU), mSTIoU$_{50}$, mSTIoU$_{75}$, \emph{mean temporal IoU} (mTIoU), mTIoU$_{50}$, mTIoU$_{75}$, \emph{Recovery} (Rec), and \emph{Success} (Succ). mSTIoU measures spatio-temporal prediction accuracy based on the mask IoU between predicted and ground-truth spatio-temporal masks, with mSTIoU$_{50}$ and mSTIoU$_{75}$ counting predictions as correct only when the IoU exceeds 0.5 and 0.75, respectively. mTIoU evaluates temporal localization accuracy using the temporal IoU between predicted and ground-truth object occurrences, with mTIoU$_{50}$ and mTIoU$_{75}$ defined analogously. Following~\cite{grauman2022ego4d}, the recovery is to evaluate the percentage of predicted frames in which the mask IoU between mask prediction and groundtruth exceeds 0.5, and the success measures weather the mask IoU between prediction and groundtruth is above 0.2. We use a stricter Success threshold of 0.2, instead of 0.05 as in~\cite{grauman2022ego4d}, because VQS-4K videos are shorter and contain more frequent object occurrences. Due to limited space, the detailed calculation of the adopted metrics is provided in the \emph{appendix}.


\section{Methodology: A New Baseline for VQS}

\textbf{Formulation.} VQS is formulated as a spatio-temporal segmentation task. Given an untrimmed video and a visual query with a mask indicating the object of interest, the goal is to segment \emph{all pixel-level} target occurrences. The results is a spatial-temporal mask response $\mathcal{R}=\{R_{i}\}_{i=1}^{P}$ containing one or multiple masklets of object occurrences, where $P$ denotes the number of object occurrences and each $R_i=\{r_{i}^{s},\cdots,r_{i}^{e}\}$ represents the $i^{\text{th}}$ object occurrence with $e$ and $s$ its start and end frame indices and $r_{i}^{j}$ the mask indicating the target in frame $j$. Examples of $R$ can be seen in Fig.~\ref{fig:task} (b).

\begin{figure}[!t]
    \centering
    \includegraphics[width=0.95\linewidth]{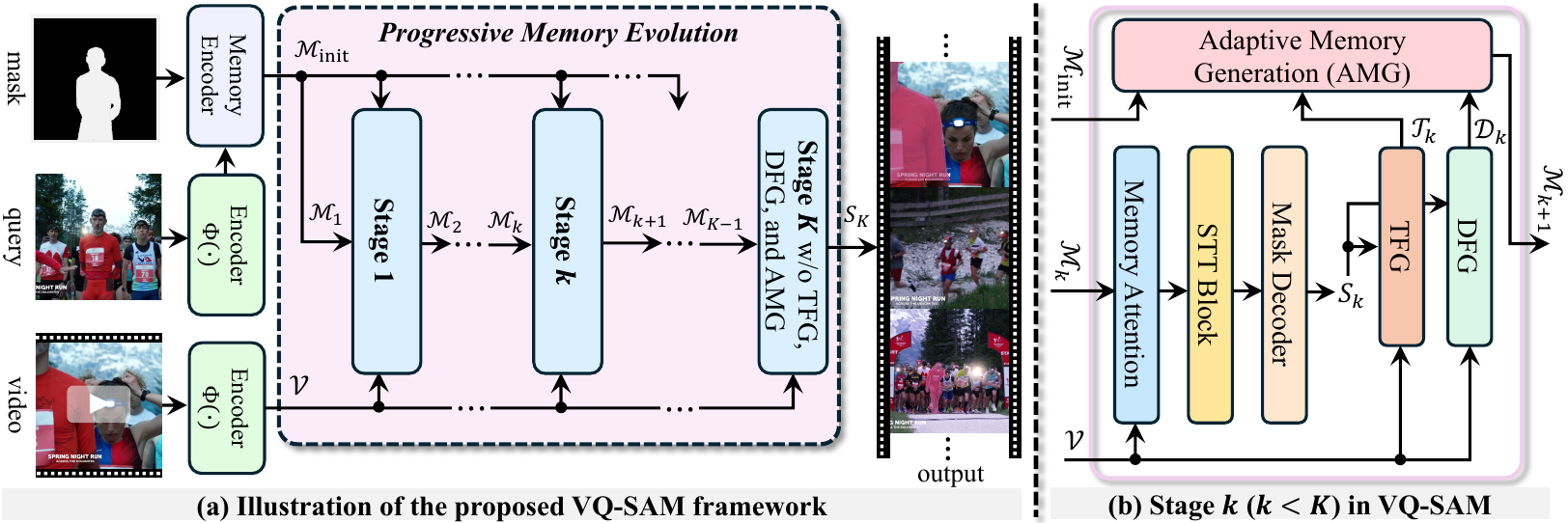}
    \caption{Overview of VQ-SAM that mines target and background distractor cues for VQS. Please \emph{\textbf{note}}, the last stage $K$ outputs $S_{K}$ for the final prediction, and TFG, DFG, and AMG are removed.}
    \vspace{-1em}
    \label{fig:framework}
\end{figure}

\textbf{Overview.} We propose VQ-SAM for VQS by extending SAM 2~\cite{RaviGHHR0KRRGMP25}. The core is to exploit target and distractor cues via a multi-stage framework to progressively evolve the memory for VQS (Sec.~\ref{vqsam}). As in Fig.~\ref{fig:framework}, at each stage, except for the final prediction stage, VQ-SAM mines confident target and distractor features using current memory. These features, together with the initial memory from visual query, are sent to AMG (Sec.~\ref{amg}) to generate new memory for the next stage, enabling more accurate target and distractor features for further memory evolution. Through this progressive process, the memory in VQ-SAM is gradually improved, leading to better segmentation in the final stage.

\subsection{The VQ-SAM Framework}
\label{vqsam}

\textbf{Feature Extraction.} In VQ-SAM, we first extract features for the visual query and video frames. Given visual query $Q$ and $L$ frames $\mathcal{I}=\{I_i\}_{i=1}^{L}$ of the video, we use a shared encoder $\mathrm{\Phi}(\cdot)$ to extract their features $f_{Q}=\mathrm{\Phi}(Q)$ and $\mathcal{V}=\{f_{i}\}_{i=1}^{L}$ with each $f_{i}=\mathrm{\Phi}(I_i)$. Afterwards, $f_{Q}$, together with the given target mask $m_{Q}$, are applied to generate the initial memory $\mathcal{M}_{\text{init}}=\texttt{MemEnc}(f_{Q},m_{Q})$, where $\texttt{MemEnc}(\cdot,\cdot)$ is memory encoder in SAM 2~\cite{RaviGHHR0KRRGMP25}. After feature extraction, we propose a multi-stage framework to progressively evolve the memory by exploring target and distractor cues.


\textbf{Progressive Memory Evolution for VQS.} Due to severe target appearance changes and similar distractors in videos, relying solely on the initial memory, generated by the visual query, is \emph{insufficient} for VQS. Addressing this, VQ-SAM mines target-specific and background distractor features in the video to progressively evolve the memory for VQS with a multi-stage framework. In each but the last stage, we mine the confident target and background distractor features using current memory, which, together with the initial memory, are used to generate a new memory for the next stage.

Specifically, for the $k^{\text{th}}$ $(1 \le k < K)$ stage of VQ-SAM, let $M_k$ be the current memory. In the first stage ($k=1$), $\mathcal{M}_1=\mathcal{M}_{\text{init}}$. To mine target and distractor features, we first generate mask candidates in each frame using $\mathcal{M}_k$, and then select target and distractor masks from them to extract features. Concretely, we first fuse current memory $\mathcal{M}_k$ into the video features using memory attention via
\begin{equation}\label{eq1}
\setlength{\abovedisplayskip}{3pt}
\setlength{\belowdisplayskip}{3pt}
    \Tilde{\mathcal{V}}_{k}=\{\Tilde{f}_{i}^{k}\}_{i=1}^{L} \;\;\;\;\; \Tilde{f}_{i}^{k} = \texttt{MemAtt}(f_{i}, \mathcal{M}_k)
\end{equation}
where $\Tilde{\mathcal{V}}_{k}$ denotes the video features after fusion, and $\texttt{MemAtt}(\cdot,\cdot)$ is the memory attention~\cite{RaviGHHR0KRRGMP25}. To capture video temporal context, we employ a spatial-temporal Transformer (STT) block to enhance video features before mask decoding via 
\begin{equation}\label{eq2}
    \setlength{\abovedisplayskip}{3pt}
    \setlength{\belowdisplayskip}{3pt}
    \Hat{\mathcal{V}}_{k}=\texttt{STTB}(\Tilde{\mathcal{V}}_k)
\end{equation}
where $\Hat{\mathcal{V}}_{k}=\{\Hat{f}_{i}^{k}\}_{i=1}^{L}$ denotes the enhanced video features with $\Hat{f}_{i}^{k}$ the feature of frame $i$. $\texttt{STTB}(\cdot)$ is the SST block, composed of feature flattening, self-attention layers~\cite{VaswaniSPUJGKP17}, and feature reshaping. After this, we apply mask decoder $\texttt{MaskDec}(\cdot)$ in~\cite{RaviGHHR0KRRGMP25} on each $\Hat{f}_{i}^{k}$ to obtain mask candidates in frame $i$ via
\begin{equation}\label{eq3}
\setlength{\abovedisplayskip}{3pt}
\setlength{\belowdisplayskip}{3pt}
    M_{i}^{k}=\texttt{MaskDec}(\Hat{f}_{i}^{k})
\end{equation}
where $M_{i}^{k}=\{m_{i,h}^k\}_{h=1}^{H}$ denotes the set of $H$ mask candidates ($H$ is set to 3 as in~\cite{RaviGHHR0KRRGMP25}). In $M_{i}^{k}$, each predicted $m_{i,h}^k$ contains the mask token, a predicted IoU score, and an occlusion score.

Once obtaining mask candidates $S_k=\{M_{i}^k\}_{i=1}^{L}$ in the video, we employ two modules, \ie, target feature generation (TFG) and distractor feature generation (DFG) as explained later, to select the target-specific and distractor masks and extract their features, as follows,
\begin{equation}\label{eq4}
\setlength{\abovedisplayskip}{3pt}
\setlength{\belowdisplayskip}{3pt}
    \mathcal{T}_k = \texttt{TFG}(\mathcal{V},S_k) \;\;\;\;\; \mathcal{D}_k =\texttt{DFG}(\mathcal{V},S_k)
\end{equation}
where $\mathcal{T}_k$ and $\mathcal{D}_k$ are the features of selected $n_t$ target and $n_d$ distractor masks, which, together the initial memory $\mathcal{M}_{\text{init}}$ from query, are fed to AMG (Sec~\ref{amg}) to generate the new memory, as follows,
\begin{equation}\label{eq5}
\setlength{\abovedisplayskip}{3pt}
\setlength{\belowdisplayskip}{3pt}
    \mathcal{M}_{k+1} = \texttt{AMG}(\mathcal{M}_{\text{init}}, \mathcal{T}_k, \mathcal{D}_k)
\end{equation}
where $\mathcal{M}_{k+1}$ is the new memory, which is sent to the next stage $(k+1)$ to generate more accurate target and distractor features for further memory evolution.

In the final $K^{\text{th}}$ stage, we perform localization and segmentation, and TFG, DFG, and AMG modules are removed. Given $\mathcal{M}_{K}$ and video features $\mathcal{V}$, we first generate the mask candidates $S_{K}=\{M_{i}^{K}\}_{i=1}^{L}$ via Eqs.~(\ref{eq1})-(\ref{eq3}), and then identify the mask with the highest IoU score in each frame $i$. This mask is selected as the final prediction in frame $i$ if its corresponding occlusion score is above 0; otherwise, the prediction is set to empty \ie, target does not exist.

\textbf{Target Feature Generation (TFG).} TFG aims to select target masks from a video and extract their features. Given $S_k=\{M_{i}^k\}_{i=1}^{L}$, we first identify mask $b_i^k \in M_{i}^k$ which has the highest IoU score in each frame $i$. To ensure mask quality, we remove those $b_i^k$ whose IoU scores are below a threshold $\tau_{\text{t}}$. After this, from the remaining masks, we select the top $n_t$ ones ranked by IoU score, which are then encoded using the memory encoder in~\cite{RaviGHHR0KRRGMP25} to obtain target features $\mathcal{T}_k$. Since $\mathcal{T}_k$ is derived from reliable target-specific masks, it can be used to improve memory to handle target appearance changes. 

\textbf{Distractor Feature Generation (DFG).} Unlike TFG, DFG selects a few distractor masks and extract their features. Similar to~\cite{videnovic2025distractor}, the distractor masks are selected from alternative mask candidates, \ie, all candidates except the best one in each frame. We first identify the best mask $b_i^k \in M_{i}^k$ in frame $i$ with the highest IoU score. Then, for each alternative mask $x$, we compute its divergence score with respect to $b_i^k$ via $1-\texttt{IoU}(b_i^k, x)$, where $\texttt{IoU}()$ computes the mask IoU, to measure spatial discrepancy between $b_i^k$ and $x$. To ensure diversity from the target and discriminative difficulty, we only keep alternative masks whose divergence and predicted IoU scores exceed thresholds $\tau_d$ and $\tau_s$. From all qualified alternative masks, we select the top $n_d$ masks ranked by the product of divergence and IoU scores, and extract their features as in~\cite{videnovic2025distractor} to obtain the distractor features $\mathcal{D}_k$. If there are no qualified alternative masks, $\mathcal{D}_k$ is empty and ignored in generating new memory. By incorporating $\mathcal{D}_k$ into memory, VQ-SAM is able to distinguish target from background distractors in localization.

\subsection{Adaptive Memory Generation (AMG)}
\label{amg}

\setlength{\columnsep}{10pt}%
\setlength\intextsep{0pt}
\begin{wrapfigure}{r}{0.5\textwidth}
\centering
\includegraphics[width=0.5\textwidth]{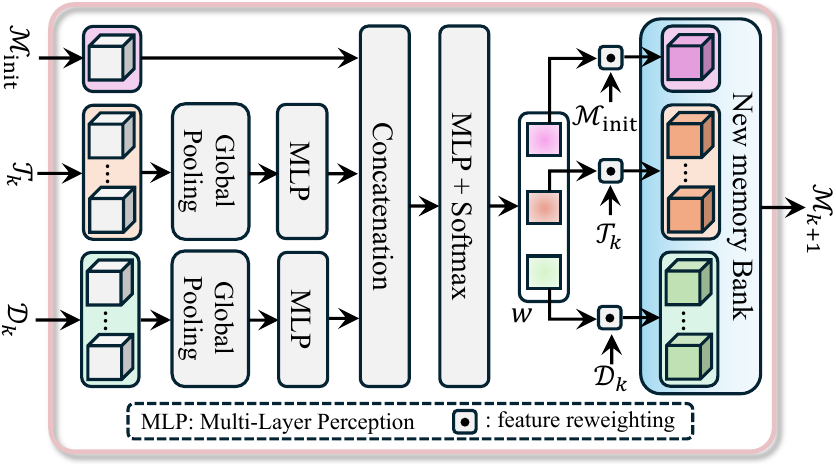}
\caption{Architecture of our AMG.}
\vspace{12pt}
\label{fig:amg}
\end{wrapfigure}
Considering different features, including $\mathcal{M}_{\text{init}}$, $\mathcal{T}_{k}$, and $\mathcal{D}_{k}$, may contribute unequally in the new memory, we design an adaptive memory generation (AMG) module that dynamically learns their relative importance and integrates them accordingly. Specifically, we first apply global pooling to $\mathcal{T}_{k}$ and $\mathcal{D}_{k}$, respectively, to fuse target and distractor features, followed by an MLP for dimensionality reduction, as follows,
\begin{equation}\label{eq6}
\setlength{\abovedisplayskip}{3pt}
\setlength{\belowdisplayskip}{3pt}
    \tilde{\mathcal{T}}_{k} = \texttt{MLP}(\texttt{GP}(\mathcal{T}_{k})) \;\; \tilde{\mathcal{D}}_{k} = \texttt{MLP}(\texttt{GP}(\mathcal{D}_{k}))
\end{equation}
Then, we concatenate $\mathcal{M}_{\text{init}}$, $\tilde{\mathcal{T}}_{k}$, and $\tilde{\mathcal{D}}_{k}$, and feed them into an MLP followed by softmax normalization to learn adaptive importance weights via
\begin{equation}
\setlength{\abovedisplayskip}{3pt}
\setlength{\belowdisplayskip}{3pt}
    w=\texttt{Softmax}(\texttt{MLP}(\texttt{Concat}(\mathcal{M}_{\text{init}}, \tilde{\mathcal{T}}_{k},\tilde{\mathcal{D}}_{k})))
\end{equation}
where $w=[w_1, w_2, w_3]\in \mathbb{R}^{1\times 3}$ represents the learned weights for $\mathcal{M}_{\text{init}}$, $\mathcal{T}_{k}$, and $\mathcal{D}_{k}$, respectively. Afterwards, we generate the new memory $\mathcal{M}_{k+1}$, as follows,
\begin{equation}
\setlength{\abovedisplayskip}{3pt}
\setlength{\belowdisplayskip}{3pt}
    \mathcal{M}_{k+1} = \{w_1 \cdot \mathcal{M}_{\text{init}}, w_2 \cdot \mathcal{T}_{k}, w_3 \cdot \mathcal{D}_{k}\}
\end{equation}

Fig.~\ref{fig:amg} shows the architecture of AMG.  Please note, when $\mathcal{D}_{k}$ is empty due to no distractors, we only learn the weights $w=[w_1,w_2]\in \mathbb{R}^{1\times 2}$ for $\mathcal{M}_{\text{init}}$ and $\mathcal{T}_{k}$. Compared with memory generation using fixed or uniform weights, AMG adaptively modulates the contribution of different feature sources, enabling more context-aware and discriminative memory generation.

\subsection{Optimization}

In VQ-SAM, we predict mask candidates $S_k=\{M_i^k\}_{i=1}^{L}$ for each stage $k$ for loss optimization. Due to limited space, please see the detailed formulation of our loss function in the \emph{appendix}.



\section{Experiments}


\textbf{Implementation.} VQ-SAM is implemented using PyTorch~\cite{paszke2019pytorch}. The encoder $\mathrm{\Phi}(\cdot)$ adopts Hiera-B+~\cite{ryali2023hiera} with feature pyramid network (FPN)~\cite{lin2017feature} for feature extraction. We train VQ-SAM for 144 epoches using AdamW~\cite{LoshchilovH19} with an initial learning of 5$\times 10^{-6}$. During training and inference, the query image and video frames are resized to 768$\times$768. The video length $L$ is set to 7 due to memory constraints. The number of stages $K$ is 2. The thresholds $\tau_t$, $\tau_d$, and $\tau_s$ are set to 0.5, 0.5, and 0.7. The numbers of selected target and distractor masks $n_t$ and $n_d$ in TFG and DFG are set to 2 and 1. 

\subsection{State-of-the-art Comparison}

Since there are no available approaches for VQS, we adapt VOS models~\cite{cheng2024putting,qin2025structure,RaviGHHR0KRRGMP25,ding2025sam2long,lee2025lomm,videnovic2025distractor,carion2025sam} and VQL approaches~\cite{jiang2023single,fan2025prvql,khosla2025ren} for comparison. For VOS methods, the target reference is the visual query with object mask. For VQL models, an additional prediction head with convolutional and upsampling layers is added to output all target masks in a video. For fair comparison, all these models, except for the train-free method~\cite{khosla2025ren}, are trained on VQS-4K$_\text{Tra}$. 

\renewcommand{\arraystretch}{1.05}
\begin{table}[!t]
\setlength{\tabcolsep}{3pt}
  \centering
  \caption{Comparison with current approaches on VQS-4K$_\text{Tst}$.}\vspace{1mm}
  \resizebox{0.97\textwidth}{!}{
    \begin{tabular}{rccccccccc}
    \Xhline{1.2pt}
    Methods & mSTIoU & mSTIoU$_{50}$ & mSTIoU$_{75}$ & mTIoU & mTIoU$_{50}$ & mTIoU$_{75}$ & Rec & Succ & FPS \\
    \hline\hline
    Cutie-b\textcolor{gray}{\scriptsize{[CVPR'24]}}~\cite{cheng2024putting} & 18.5  & 16.5  & 8.2   & 21.0  & 16.9  & 9.2   & 38.4  & 32.3 & 36 \\
    OASIS\textcolor{gray}{\scriptsize{[ICCV'25]}}~\cite{qin2025structure} & 15.8  & 13.6  & 6.7   & 21.3  & 17.4  & 9.5   & 34.3  & 29.4 & 48 \\
    LOMM\textcolor{gray}{\scriptsize{[ICCV'25]}}~\cite{lee2025lomm} & 12.5  & 10.2  & 5.0   & 12.7  & 7.9   & 4.3   & 26.1  & 25.8 & 10 \\
    SAM 2\textcolor{gray}{\scriptsize{[ICLR'25]}}~\cite{RaviGHHR0KRRGMP25} & 12.9  & 10.6  & 5.2   & 13.0  & 8.3   & 4.6   & 26.7  & 26.3 & 22 \\
    SAM2.1++\textcolor{gray}{\scriptsize{[CVPR'25]}}~\cite{videnovic2025distractor} &  12.5  &  10.2  &  5.1   &  14.6  &  10.0  &  5.5  &  28.8  &  25.8 & 16 \\
    SAM2Long\textcolor{gray}{\scriptsize{[ICCV'25]}}~\cite{ding2025sam2long} & 18.6  & 16.5  & 8.2   & 24.4  & 20.6  & 11.2  & 38.6  & 32.5 & 19 \\
    SAM 3\textcolor{gray}{\scriptsize{[ICLR'26]}}~\cite{carion2025sam} & 13.3  & 10.9  & 5.4   & 14.1  & 9.5   & 5.2   & 27.7  & 26.6 & 15 \\
    VQLoC\textcolor{gray}{\scriptsize{[NeurIPS'23]}}~\cite{jiang2023single} & 10.1  & 12.5  & 6.3   & 21.5  & 17.5  & 8.4   & 31.7  & 23.2 & 36 \\
    PRVQL\textcolor{gray}{\scriptsize{[ICCV'25]}}~\cite{fan2025prvql} & 13.6  & 14.4  & 7.1   & 22.1  & 18.4  & 8.7   & 32.6  & 26.9 & 30 \\
    REN\textcolor{gray}{\scriptsize{[NeurIPS'25]}}~\cite{khosla2025ren} & 16.8  & 17.7  & 8.9   & 23.6  & 19.8  & 9.5   & 34.8  & 30.5 & 12 \\
    \hline
    \rowcolor[HTML]{E7E7FF}
    VQ-SAM (ours) & 26.0  & 23.8  & 12.7  & 29.6  & 26.8  & 14.4  & 43.6  & 42.1 & 29 \\
    \Xhline{1.2pt}
    \end{tabular}}
  \label{tab:sotacomp}%
  \vspace{-1em}
\end{table}%

\renewcommand{\arraystretch}{1.05}
\begin{table}[!t]
  \centering
  \setlength{\tabcolsep}{2pt}
  \caption{Comparison on VQS-4K$_{\text{Tst}}$-Small, VQS-4K$_{\text{Tst}}$-Medium, and VQS-4K$_{\text{Tst}}$-Large.}\vspace{1mm}
  \resizebox{0.97\textwidth}{!}{
    \begin{tabular}{rcccccccccccc}
    \Xhline{1.2pt}
          & \multicolumn{4}{c}{VQS-4K$_\text{Tst}$-Small}     & \multicolumn{4}{c}{VQS-4K$_\text{Tst}$-Medium}    & \multicolumn{4}{c}{VQS-4K$_\text{Tst}$-Large} \\
          \cmidrule(lr){2-5} \cmidrule(lr){6-9} \cmidrule(lr){10-13}
    Methods & mSTIoU  & mTIoU   & Rec & Succ  & mSTIoU  & mTIoU   & Rec & Succ  & mSTIoU  & mTIoU   & Rec & Succ \\
    \hline\hline
    Cutie-b\textcolor{gray}{\scriptsize{[CVPR'24]}}~\cite{cheng2024putting} & 13.3  & 17.2  & 29.9  & 25.5  & 19.0  & 21.4  & 39.6  & 33.8  & 25.6  & 26.1  & 45.4  & 39.8 \\
    OASIS\textcolor{gray}{\scriptsize{[ICCV'25]}}~\cite{qin2025structure} & 11.4  & 17.4  & 26.7  & 23.2  & 16.2  & 21.7  & 35.4  & 30.7  & 21.9  & 26.5  & 40.5  & 36.2 \\
    LOMM\textcolor{gray}{\scriptsize{[ICCV'25]}}~\cite{lee2025lomm}  & 9.0   & 10.4  & 20.4  & 20.3  & 12.8  & 13.0  & 26.9  & 27.0  & 17.3  & 15.8  & 30.8  & 31.8 \\
    SAM 2\textcolor{gray}{\scriptsize{[ICLR'25]}}~\cite{RaviGHHR0KRRGMP25} & 9.3   & 10.6  & 20.8  & 20.7  & 13.2  & 13.3  & 27.6  & 27.5  & 17.9  & 16.2  & 31.6  & 32.4 \\
    SAM2.1++\textcolor{gray}{\scriptsize{[CVPR'25]}}~\cite{videnovic2025distractor} & 9.1   & 11.9  & 22.5  & 20.3  & 12.8  & 14.9  & 29.7  & 27.0  & 17.3  & 18.1  & 34.0  & 31.8 \\
    SAM2Long\textcolor{gray}{\scriptsize{[ICCV'25]}}~\cite{ding2025sam2long} & 13.4  & 19.9  & 30.1  & 25.6  & 19.1  & 24.9  & 39.9  & 34.0  & 25.8  & 30.4  & 45.6  & 40.0 \\
    SAM 3\textcolor{gray}{\scriptsize{[ICLR'26]}}~\cite{carion2025sam} & 9.6   & 11.5  & 21.5  & 21.0  & 13.6  & 14.4  & 28.6  & 27.8  & 18.4  & 17.5  & 32.7  & 32.7 \\
    VQLoC\textcolor{gray}{\scriptsize{[NeurIPS'23]}}~\cite{jiang2023single} & 7.3   & 17.6  & 24.7  & 18.3  & 10.4  & 21.9  & 32.7  & 24.3  & 14.0  & 26.7  & 37.5  & 28.6 \\
    PRVQL\textcolor{gray}{\scriptsize{[ICCV'25]}}~\cite{fan2025prvql} & 9.8   & 18.1  & 25.4  & 21.2  & 14.0  & 22.5  & 33.7  & 28.1  & 18.9  & 27.5  & 38.5  & 33.1 \\
    REN\textcolor{gray}{\scriptsize{[NeurIPS'25]}}~\cite{khosla2025ren}   & 12.1  & 19.3  & 27.1  & 24.0  & 17.2  & 24.1  & 35.9  & 31.9  & 23.3  & 29.4  & 41.2  & 37.5 \\
    \hline
    \rowcolor[HTML]{E7E7FF}
    VQ-SAM (ours) & 18.8  & 24.3  & 34.0  & 33.2  & 26.7  & 30.2  & 45.0  & 44.0  & 36.0  & 36.8  & 51.5  & 51.8 \\
    \Xhline{1.2pt}
    \end{tabular}}
  \label{tab:scaleres}%
  \vspace{-1em}
\end{table}%

Tab.~\ref{tab:sotacomp} reports the results. As shown, VQ-SAM significantly outperforms other methods on all metrics. Specifically, VQ-SAM achieves the best mSTIoU score with 26.0\% and mTIoU score with 26.8\%, which surpasses the second best SAM2Long with 18.6\% mSTIoU and 24.4\% mTIoU by 7.4\% and 5.2\%, clearly showing its superiority. Compared with the VQL method REN with 16.8\% mSTIoU and 23.6\% mTIoU, VQ-SAM outperforms it by 9.2\% in mSTIoU and 6.0\% in mTIoU, evidencing the strength of our method. Compared to SAM 2, VQ-SAM obtains gains of 13.1\% and 16.6\% in mSTIoU and mTIoU, showing the importance of our memory evolution for VQS. Moreover, as shown in Tab.~\ref{tab:sotacomp}, several recent VOS methods such as OASIS, LOMM, SAM2.1++, and SAM 3 struggle in VQS. This highlights the unique challenges posed by VQS, particularly in conducting video-level target search in untrimmed videos. 

In addition, we compare different approaches across different scales on three subsets of VQS-4K, including 4K$_{\text{Tst}}$-Small, VQS-4K$_{\text{Tst}}$-Medium, and VQS-4K$_{\text{Tst}}$-Large. The results are shown in Tab.~\ref{tab:scaleres}. As shown, our VQ-SAM consistently shows the best performance for objects with different scales.

\subsection{Ablation Studies}

We conduct ablation studies on VQS-4K$_\text{Tst}$, with our final configuration highlighted in {\color{cyan!75} cyan}.

\setlength{\columnsep}{6pt}%
\setlength\intextsep{-0pt}
\begin{wraptable}{r}{0.47\textwidth}
\setlength{\tabcolsep}{6pt}
	\centering
	\renewcommand{\arraystretch}{1.05}
    \caption{Ablation on TFG and DFG.}\vspace{-3pt}
	\resizebox{0.43\textwidth}{!}{
    \begin{tabular}{ccccccc}
        \specialrule{1.5pt}{0pt}{0pt} 
         & TFG & DFG & mSTIoU & mTIoU & Rec &  Succ \\
        \hline
        \hline
         \ding{182} &  & & 23.8 & 24.9 & 36.7 & 27.1 \\
         \ding{183} & \checkmark & & 24.8 & 25.4   & 38.3 & 31.8 \\
         \ding{184} & & \checkmark  & 25.4 & 27.3 & 41.1 & 37.0 \\
         \rowcolor{cyan!10} 
         \ding{185} & \checkmark & \checkmark  & 26.0 & 29.6 & 43.6 & 42.1 \\
        \specialrule{1.5pt}{0pt}{0pt}
    \end{tabular}}
    \label{tab:connect}
 \vspace{3pt}
\end{wraptable}
\textbf{Impact of TFG and DFG.} To analyze the impacts of TFG and DFG, we conduct ablation studies in Tab.~\ref{tab:connect}. As displayed in Tab.~\ref{tab:connect}, without TFG and DFG, the mSTIoU and mTIoU scores are 23.8\% and 24.9\% (\ding{182}). When applying TFG along, mSTIoU and mTIoU are improved to 24.8\% and 25.4\% (\ding{183}), showing that target information from the video benefits target segmentation. When using DFG along, mSTIoU and mTIoU are improved to 25.4\% and 27.3\% (\ding{184}), indicating the importance of exploring distractor cues for VQS. When using both TFG and DFG, we achieve the best results with 26.0\% mSTIoU and 29.6\% mTIoU (\ding{185}), evidencing the efficacy of memory evolution for VQS. 

\setlength{\columnsep}{6pt}%
\setlength\intextsep{-0pt}
\begin{wraptable}{r}{0.47\textwidth}
\setlength{\tabcolsep}{8pt}
	\centering
	\renewcommand{\arraystretch}{1.05}
    \caption{Ablation on memory generation.}\vspace{-3pt}
	\resizebox{0.43\textwidth}{!}{
    \begin{tabular}{cccccc}
        \specialrule{1.5pt}{0pt}{0pt} 
         & & mSTIoU & mTIoU & Rec & Succ \\
        \hline
        \hline
         \ding{182} & EMG & 23.9 & 27.0 & 41.2 & 38.4 \\
         \ding{183} & SLMG & 25.1 & 28.7 & 42.4 & 39.7  \\
         \rowcolor{cyan!10}
         \ding{184} & AMG & 26.0 & 29.6 & 43.6 & 42.1 \\
        \specialrule{1.5pt}{0pt}{0pt}
    \end{tabular}}
    \label{tab:amg}
 \vspace{3pt}
\end{wraptable}
\textbf{Impact of AMG.} In VQ-SAM, we propose AMG to integrate different features for memory generation. To study its effectiveness, we conduct an ablation study by comparing AMG with two alternative strategies: equal-weight memory generation (EMG) and static learnable memory generation (SLMG). EMG integrates different features using fixed equal weights, whereas SLMG assigns learnable weights to each feature type, which are fixed after training. As shown in Tab.~\ref{tab:amg}, using AMG achieves the best results (\ding{184} \emph{v.s} \ding{182}/\ding{183}), showing its efficacy.

\setlength{\columnsep}{6pt}%
\setlength\intextsep{-0pt}
\begin{wraptable}{r}{0.47\textwidth}
\setlength{\tabcolsep}{6pt}
	\centering
	\renewcommand{\arraystretch}{1.05}
    \caption{Ablation on the STT block.}\vspace{-3pt}
	\resizebox{0.43\textwidth}{!}{
    \begin{tabular}{cccccc}
        \specialrule{1.5pt}{0pt}{0pt} 
         & & mSTIoU & mTIoU & Rec &  Succ \\
        \hline
        \hline
         \ding{182} & w/o STT block & 25.2 & 27.9 & 39.7 & 36.1 \\
         \rowcolor{cyan!10}
         \ding{183} & w/ STT block & 26.0 & 29.6 & 43.6 & 42.1  \\
        \specialrule{1.5pt}{0pt}{0pt}
    \end{tabular}}
    \label{tab:stt}
 \vspace{3pt}
\end{wraptable}
\textbf{Impact of the STT block.} To capture global temporal context in the video for VQS, we apply STT in VQ-SAM. To investigate its impact, we conduct ablation studies in Tab.~\ref{tab:stt}. As shown, without STT, the mSTIoU and mTIoU are 25.2\% and 27.9\% (\ding{182}). When using STT, mSTIoU and mTIoU are improved to 26.0\% and 29.6\% (\ding{183}), showing its efficacy.

\setlength{\columnsep}{6pt}%
\setlength\intextsep{-0pt}
\begin{wraptable}{r}{0.47\textwidth}
\setlength{\tabcolsep}{6pt}
	\centering
	\renewcommand{\arraystretch}{1.05}
    \caption{Ablation on number of stages.}\vspace{-3pt}
	\resizebox{0.43\textwidth}{!}{
    \begin{tabular}{cccccc}
        \specialrule{1.5pt}{0pt}{0pt} 
         &  & mSTIoU & mTIoU & Rec &  Succ \\
        \hline
        \hline
         \ding{182} & $K=1$ & 21.5 & 13.9 & 17.3 & 8.2 \\
         \rowcolor{cyan!10}
         \ding{183} & $K=2$ & 26.0 & 29.6 & 43.6 & 42.1 \\
         \ding{184} & $K=3$ & 25.1 & 28.6 & 44.8 & 39.8 \\
        \specialrule{1.5pt}{0pt}{0pt}
    \end{tabular}}
    \label{tab:stage}
 \vspace{3pt}
\end{wraptable}
\textbf{Impact of the number of stages.} We conduct ablation studies on the number $K$ of stages in VQ-SAM in Tab.~\ref{tab:stage}. As shown, when setting $K=2$, we achieve the best mSTIoU of 26.0\% and mTIoU of 29.6\% (\ding{183}). When further increasing the stage number to 3 (\ding{184}), the performance is slightly decreased. Thus, we set $K$ to 2 in this work.

\subsection{Comparison with Existing VQL Methods on VQ2D}

\setlength{\intextsep}{-0pt}
\setlength{\columnsep}{9pt}%
\begin{wraptable}{r}{0.47\textwidth}
\setlength{\tabcolsep}{3pt}
	\centering
	\renewcommand{\arraystretch}{1.05}
    \caption{Comparison with VQL models.}\vspace{-2pt}
	\scalebox{0.75}{
    \begin{tabular}{lcccc}
    \Xhline{1.2pt}
    Methods & mSTIoU & mTIoU & Rec & Succ \\ \hline\hline
    VQLoC\textcolor{gray}{\scriptsize{[NeurIPS'23]}}~\cite{jiang2023single} & 23.7  & 32.3  & 45.1  & 55.9 \\
    PRVQL\textcolor{gray}{\scriptsize{[ICCV'25]}}~\cite{fan2025prvql}  & 28.1  & 36.8  & 45.7  & 59.4 \\
    RELOCATE\textcolor{gray}{\scriptsize{[CVPR'25]}}~\cite{khosla2025relocate}  & 35.2  & 42.9  & 50.6  & 60.1 \\
    REN{\scriptsize{DINOv2]}}\textcolor{gray}{\scriptsize{[NeurIPS'25]}}~\cite{khosla2025ren} & 39.7   & 51.6  & 49.3  & 61.2 \\ \hline
    \rowcolor[HTML]{E7E7FF}
    VQ-SAM (ours) & 41.8 & 56.0 & 53.8 & 64.2 \\
    \Xhline{1.2pt}
    \end{tabular}}
	\label{tab:vql_res}
\end{wraptable}
Although our VQ-SAM is designed for VQS, it can also be used for VQL. To validate this, we compare it with VQL methods on VQ2D validate set~\cite{grauman2022ego4d}. Please note that, for the VQL evaluation, the mask outputs of VQ-SAM are converted into bounding boxes. Tab.~\ref{tab:vql_res} reports the results. As shown, even VQ-SAM is not specifically design for VQL, it still achieves the best results, validating its efficacy.

Due to limited space, please refer to our \emph{appendix} for more results, analysis, and discussion.

\section{Conclusion}

This paper introduces VQS, a new paradigm of VQL aiming at segmenting all pixel-level occurrences of an object from an untrimmed video. To facilitate research on VQS, we propose VQS-4K by including 4,111 videos. Moreover, to stimulate future research, we present VQ-SAM, a simple but effective method for VQS. Our experiments on VQS-4K validates the effectiveness of VQ-SAM. Through VQS-4K and VQ-SAM, we hope to inspire more research and applications on VQS.

\section*{Acknowledgments and Disclosure of Funding} 
This work receives no external funding. The authors declare no competing interests.



{
\bibliographystyle{plainnat}
\bibliography{reference}
}







\newpage
\appendix

\section*{\textbf{\Large{Appendix}}}

To better understand this work, we offer additional details, analysis, and results:

\vspace{-0.3em}
\begin{itemize}
	\setlength{\itemsep}{2pt}

   \item \textbf{S1 \; Details of Object Categories} \\
   In this section, we provide the details of object classes in VQS-4K.
   
   \item \textbf{S2 \; Construction Pipeline of VQS-4K and Additional Annotation Examples} \\
   In this section, we describe the detailed construction pipeline of our VQS-4K, and provide more annotation examples.
   
   \item \textbf{S3 \; More Statistics} \\
   This section shows more statistics on our benchmark,

   \item \textbf{S4 \; Details of Evaluation Metrics} \\
   In this section, we offer detailed calculation of adopted evaluation metrics.

   \item \textbf{S5 \; Details of Optimization} \\
   In this section, we show the details of loss function for optimization.

   \item \textbf{S6 \; Additional Results and Analysis} \\
   In this section, we provide additional results and analysis of our work.

   \item \textbf{S7 \; Ethical Statement and Maintenance of VQS-4K} \\
   This section discusses the ethical statement and maintenance of VQS-4K.

   \item \textbf{S8 \; Discussion on Limitation and Broader Impact} \\
   We discuss the limitation and broader impact of our work in this section.

\end{itemize}

\begin{figure}[h]
   \centering
   \includegraphics[width=\textwidth]{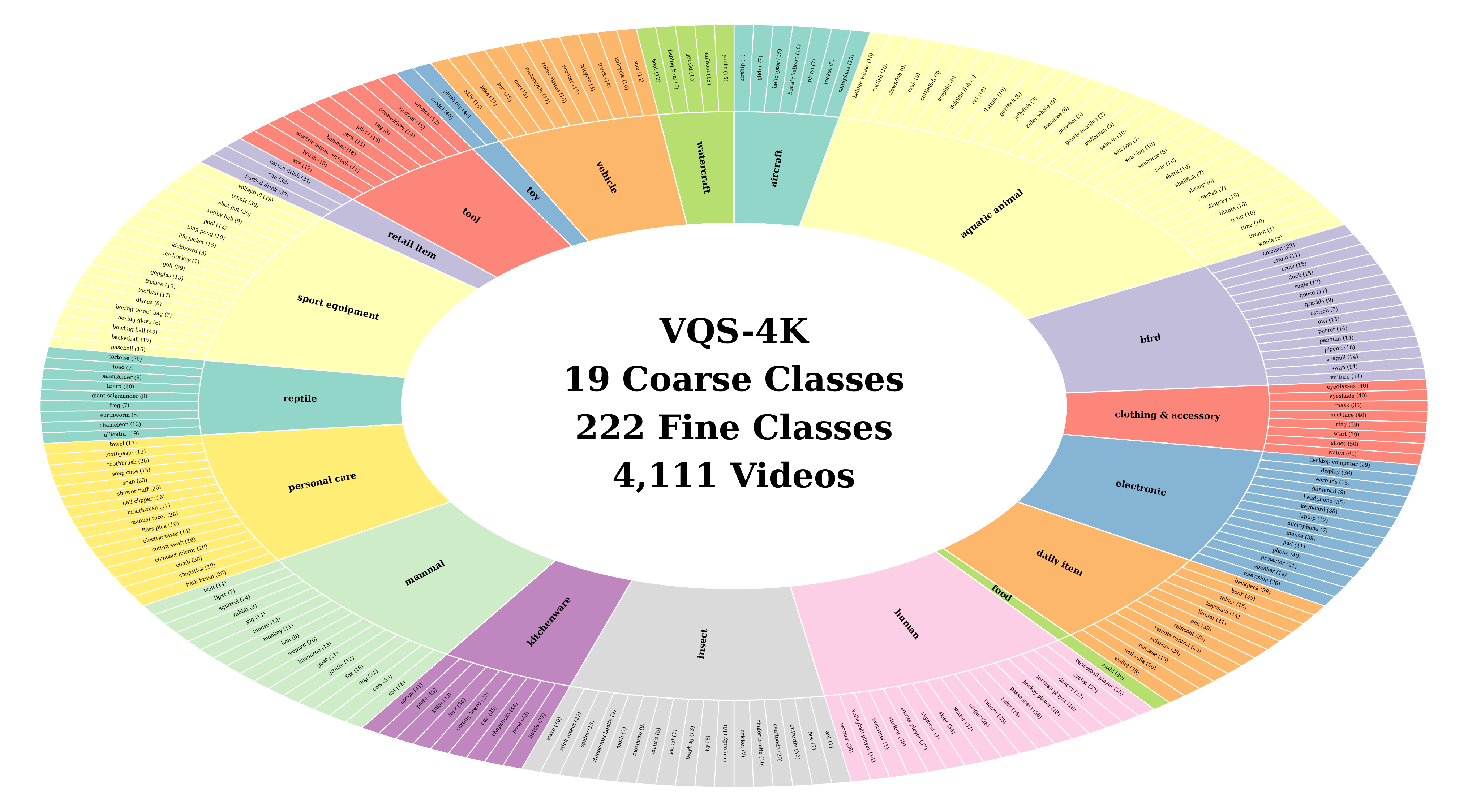}
   \caption{Category organization of VQS-4K. The inner part of chart displays 19 coarser classes, while the outer part shows 222 fine categories. Best viewed by zooming in.}
   \label{fig:class-dist}
\end{figure}

\section*{S1 \; Details of Object Categories}

VQS-4K comprises 222 object classes, aiming to offer a diverse platform for VQS. These categories are organized in the hierarchical structure. Specifically, we first determine 19 coarse object classes, including ``\emph{human}'', ``\emph{insect}'', ``\emph{kitchenware}'', ``\emph{mammal}'', ``\emph{personal care}'', ``\emph{reptile}'', ``\emph{sport equipment}'', ``\emph{retail item}'', ``\emph{tool}'', ``\emph{toy}'', ``\emph{vehicle}'', ``\emph{watercraft}'', ``\emph{aircraft}'', ``\emph{aquatic animal}'', ``\emph{bird}'', ``\emph{clothing \& accessory}'', ``\emph{electronic}'', ``\emph{daily item}'', and ``\emph{food}''. Please note that, since ``human'' is a special category, we separate it from ``mammal''. After this, we further gather 222 fine categories
from 19 coarse classes. Fig.~\ref{fig:class-dist} shows the category organization of VQS-4K (please zoom in for best view).

\begin{figure}[!t]
   \centering
   \includegraphics[width=\textwidth]{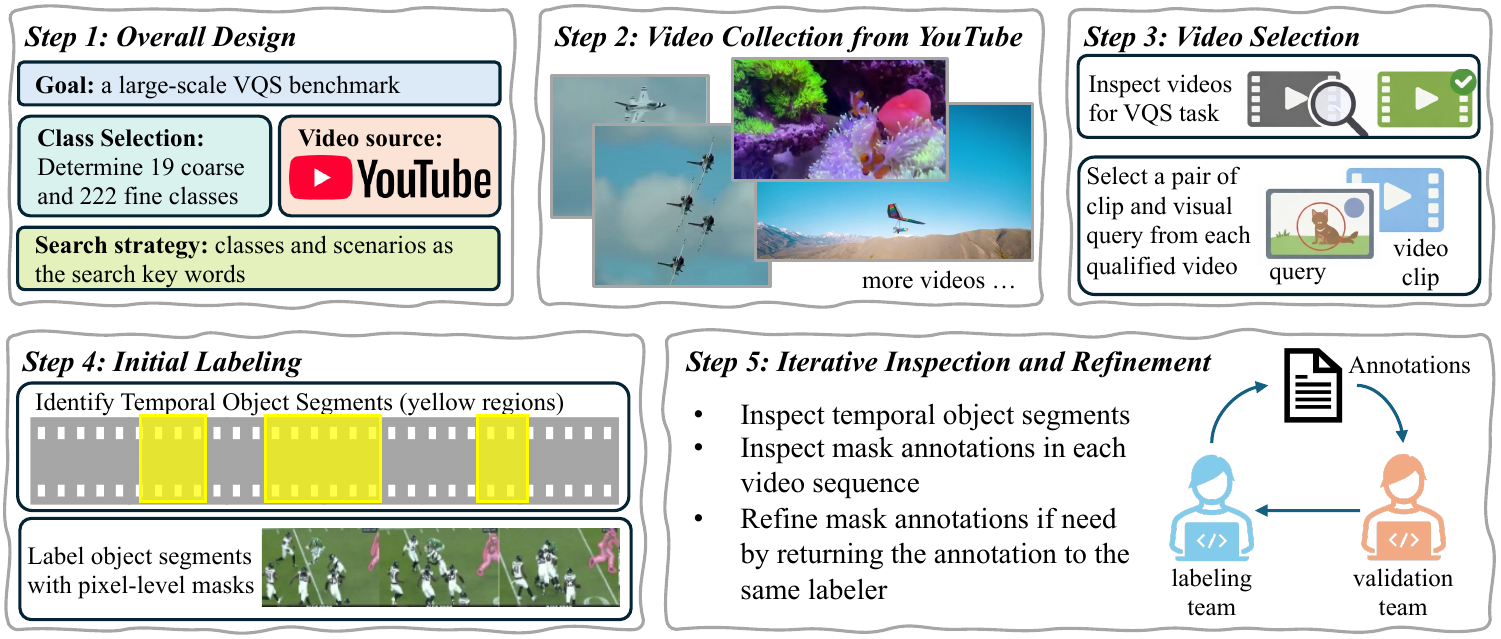}
   \caption{Construction pipeline of VQS-4K, including five steps, \ie, overall design, video collection, video selection, initial labeling, and iterative inspection and refinement.}
   \label{fig:anno_pipeline}
\end{figure}

\section*{S2 \; Construction Pipeline of VQS-4K and Additional Annotation Examples}

\begin{figure}[!t]
   \centering
   \includegraphics[width=\textwidth]{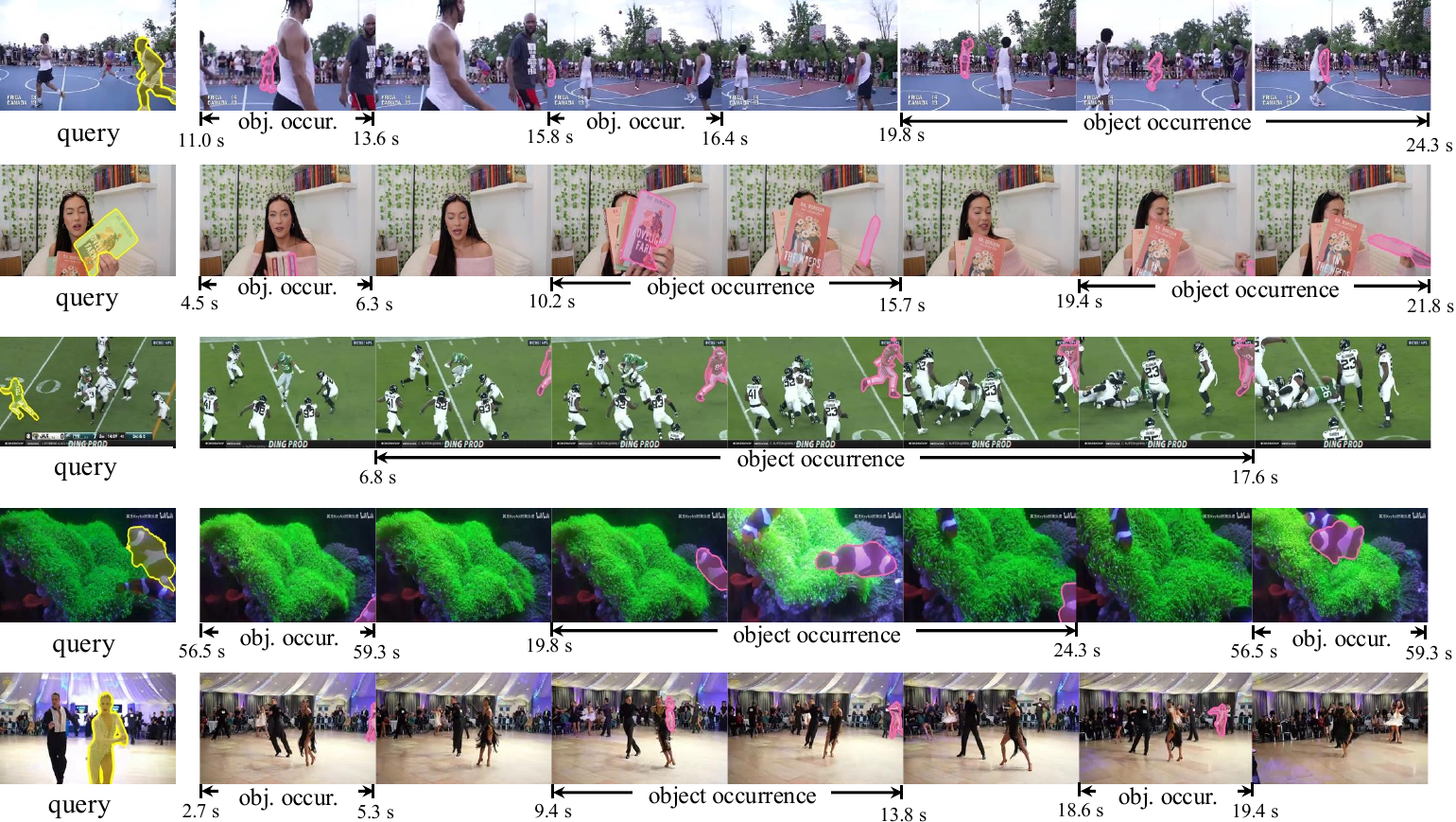}
   \caption{Additional annotation samples in the proposed VQS-4K.}
   \label{fig:addi_anno}
\end{figure}

In this paper, we adopt a systematic five-step pipeline to construct VQS-4K: \textbf{(i) \emph{Overall design}}. In the first step, we determine all object categories in VQS-4K and a video search strategy using class names and scenarios as keywords. \textbf{(ii) \emph{Video collection}}. The second step is to collect video sequences from YouTube using the strategy in the first step. Please note, all videos in VQS-4K are collected under the \emph{creative commons} license and used for \emph{research purpose only}. \textbf{(iii) \emph{Video selection}}. The third step is to inspect videos collect in the second step to verify their availabilities for VQS. We will select a pair of clip and visual query from each qualified video. \textbf{(iv) \emph{Initial labeling}}. The fourth step is to conduct initial labeling by experts. Specifically, for each pair of video and query, we first identify all temporal segments in the video in which the queried object appears. Then, within each identified temporal segment, pixel-level masks will be provided for the target of interest. \textbf{(v) \emph{Iterative inspection and refinement}}. In the final step, multiple rounds of iterative inspection and refinement if needed are performed by experts and volunteers to ensure high annotation quality. Fig.~\ref{fig:anno_pipeline} shows the construction pipeline. In Fig.~\ref{fig:addi_anno}, we provide more annotation examples.

\section*{S3 \; More Statistics}

\setlength{\columnsep}{10pt}%
\setlength\intextsep{5pt}
\begin{wrapfigure}{r}{0.6\textwidth}
\centering
\includegraphics[width=0.55\textwidth]{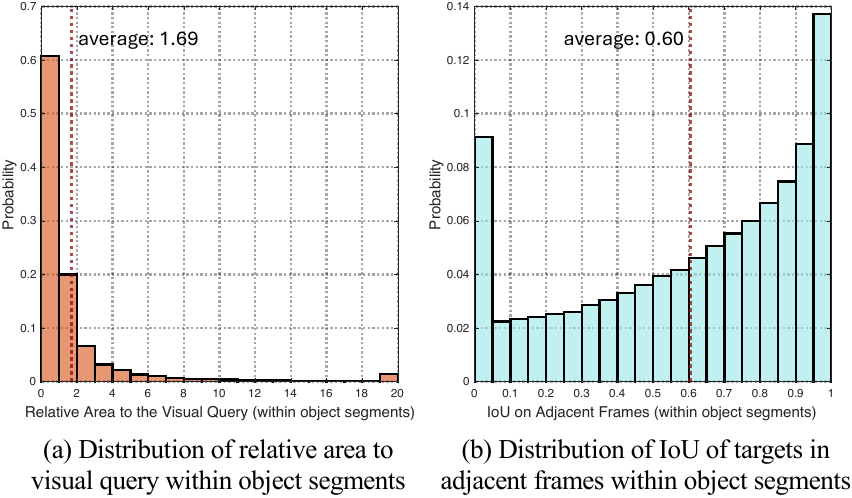}
\caption{More statistics.}\vspace{-5mm}
\label{fig:supp_plot_stat}
\end{wrapfigure}
In order to better understand our VQS-4K, we further demonstrate other representative statistics. Specifically, we show the distributions of relative area to the initial target and Intersection over Union (IoU) between the targets in adjacent frames with in temporal object segments in Fig.~\ref{fig:supp_plot_stat}. These two statistics can reflect the motion information of the target in the video. From Fig.~\ref{fig:supp_plot_stat}, we can observe that queried target moves fast and varies in scale rapidly in the videos.


\section*{S4 \; Details of Evaluation Metrics}

We adopt multiple metrics to comprehensively evaluate both temporal and spatial-temporal localization performance for VQS. We report mSTIoU, mSTIoU$_{50}$, mSTIoU$_{75}$, mTIoU, mTIoU$_{50}$, mTIoU$_{75}$, Recovery (Rec), and Success (Succ). Specifically, mSTIoU and mTIoU measure localization accuracy by averaging per-video overlap scores over the testing set VQS-4K$_{\text{Tst}}$, and are calculated as follows,
\begin{equation}
    \text{mSTIoU} = \frac{1}{|\textbf{V}_{\text{tst}}|} \sum_{v \in \textbf{V}_{\text{tst}}} \text{stIoU}_v 
    \quad 
     \text{mTIoU} = \frac{1}{|\textbf{V}_{\text{tst}}|} \sum_{v \in \textbf{V}_{\text{tst}}} \text{tIoU}_v
\end{equation}
where $\textbf{V}_{\text{tst}}$ denotes the set of videos in the VQS-4K$_{\text{Tst}}$, and $|\textbf{V}_{\text{tst}}|$ is the number of videos in $\textbf{V}_{\text{tst}}$. $\text{stIoU}_v$ and $\text{tIoU}_v$ represent the stIoU (\ie, overlap between groundtruth and predicted masks across all frames) and tIoU (\ie, overlap between ground truth and predicted temporal segments) scores for the video $v$, and are computed as follows,
\begin{equation}
    \text{stIoU}_v = \frac{\sum\limits_{t \in v_{\text{gt}} \cup v_{\text{pred}}} |\mathcal{R}_t^{\text{gt}} \cap \mathcal{R}_t^{\text{pred}}|}{\sum\limits_{t \in v_{\text{gt}}} |\mathcal{R}_t^{\text{gt}}| + \sum\limits_{t \in v_{\text{pred}}} |\mathcal{R}_t^{\text{pred}}| - \sum\limits_{t \in v_{\text{gt}} \cup v_{\text{pred}}} |\mathcal{R}_t^{\text{gt}} \cap \mathcal{R}_t^{\text{pred}}|}
\end{equation}
\begin{equation}
    \text{tIoU}_v = \frac{|\mathcal{F}_{v_\text{gt}} \cap \mathcal{F}_{v_\text{pred}}|}{|\mathcal{F}_{v_\text{gt}} \cup \mathcal{F}_{v_\text{pred}}|}
\end{equation}
where $\mathcal{R}_t^{\text{gt}}$ and $\mathcal{R}_t^{\text{pred}}$ are groundtruth and predicted masks at frame $t$ in video $v$. $\mathcal{F}_{v_\text{gt}}$ and $\mathcal{F}_{v_\text{pred}}$ are the sets of frames containing the groundtruth and predicted masks in video $v$. With mSTIoU and mTIoU, mSTIoU$_{\tau}$ and mTIoU$_{\tau}$ are computed as follows,
\begin{equation}
    \text{mSTIoU}_{\tau} = \frac{1}{|\textbf{V}_{\text{tst}}|} \sum_{v \in \textbf{V}_{\text{tst}}} \mathbb{I}[\text{stIoU}_v \geqslant \tau] \cdot \text{stIoU}_v  \quad 
    \text{mTIoU}_{\tau} = \frac{1}{|\textbf{V}_{\text{tst}}|} \sum_{v \in \textbf{V}_{\text{tst}}} \mathbb{I}[\text{tIoU}_v \geqslant \tau] \cdot \text{tIoU}_v
\end{equation}

where $\tau \in \{0.5, 0.75\}$ is the threshold. $\mathbb{I}[\cdot]$ is the indicator function that return 1 if the condition is true and 0 otherwise.

Different from mSTIoU and mTIoU, the recovery is to evaluate the percentage of predicted frames in which the mask IoU between mask prediction and groundtruth exceeds 0.5, and the success measures weather the mask IoU between prediction and groundtruth is above 0.2.

It is worth noting that, the adopted metrics differ from conventional precision-recall AP in that they do not evaluate ranked predictions or compute a precision-recall curve. Instead, they directly average the overlap between the predicted and ground-truth results across all videos. Specifically, mSTIoU is computed by averaging the mask IoU between predicted and ground-truth masks over all videos, while mTIoU is computed by averaging the IoU between predicted and ground-truth temporal segments. The thresholded variants follow the same principle by averaging the overlap scores of predictions satisfying the corresponding IoU threshold, rather than computing conventional AP at different IoU thresholds. Therefore, unlike conventional AP, which summarizes performance over different confidence thresholds based on ranked predictions, mSTIoU and mTIoU directly measure the average overlaps of the predicted results and groundtruth.

\section*{S5 \; Details of Optimization}\label{opt}

Given the video and the visual query with an object mask, VQ-SAM predicts mask candidates $S_k=\{M_i^k\}_{i=1}^{L}$ for each stage $k$ ($1 \le k \le K$; $K$ is set to 2). During training, given the groundtruth, following SAM 2~\cite{RaviGHHR0KRRGMP25}, we utilize multiple losses, including dice coefficient loss, penalties residual activation loss, IoU loss, and binary cross-entropy loss for the loss function $\mathcal{L}_k^{i}$ in the $i^{\text{th}}$ frame (please see~\cite{RaviGHHR0KRRGMP25} for details). Then, the loss $\mathcal{L}_k$ in stage $k$ is calculated as $\mathcal{L}_k=\sum_{i=1}^{L}{\mathcal{L}_k^{i}}$, and the total training loss $\mathcal{L}_{\text{total}}=\sum_{k=1}^{K}\gamma_k\mathcal{L}_k$ with $\gamma_k$ the loss weight for stage $k$. The loss weights $\gamma_1$ and $\gamma_2$ are empirically set to 0.5 and 1.0.

\section*{S6 \; Additional Results and Analysis}

\begin{table}[!t]
	\centering
	\begin{minipage}{.49\textwidth}
    \centering
    \setlength{\tabcolsep}{7pt}
    \caption{{Ablation on $n_t$ in TFG.}} \vspace{2mm}
    \label{tab:nt}
    \scalebox{0.8}{
    \begin{tabular}{cccccc}
        \specialrule{1.5pt}{0pt}{0pt} 
         &   & mSTIoU & mTIoU & Rec &  Succ \\
        \hline
        \hline
         \ding{182}  & $n_t=1$ & 25.0 & 28.6 & 42.0 & 39.8 \\
         \rowcolor{cyan!10}
         \ding{183}  & $n_t=2$ & 26.0 & 29.6 & 43.6 & 42.1  \\
         \ding{184}  & $n_t=3$ & 25.1 & 28.6	& 42.1 & 40.0  \\
        \specialrule{1.5pt}{0pt}{0pt}
    \end{tabular}}
\end{minipage}%
\hfill
\begin{minipage}{.49\textwidth}
    \centering
    \setlength{\tabcolsep}{7pt}
    \caption{{Ablation on $n_d$ in DFG.}} \vspace{2mm}
    \label{tab:nd}
    \scalebox{0.8}{
    \begin{tabular}{cccccc}
        \specialrule{1.5pt}{0pt}{0pt} 
         &  & mSTIoU & mTIoU & Rec &  Succ \\
        \hline
        \hline
        \rowcolor{cyan!10}
         \ding{182}  & $n_d=1$ & 26.0 & 29.6 & 43.6 & 42.1 \\
         \ding{183}  & $n_d=2$ & 24.2 & 28.0 & 41.3 & 38.3 \\
         \ding{184}  & $n_d=3$ & 23.7 & 26.6 & 39.2 & 37.0  \\
        \specialrule{1.5pt}{0pt}{0pt}
    \end{tabular}}
\end{minipage}%
 \vspace{-10pt}
\end{table}

\subsection*{S6.1 \; Impact of $n_t$ in TFG}

In VQ-SAM, $n_t$ controls the number of selected target masks for the new memory generation. We conduct an ablation on $n_t$ in Tab.~\ref{tab:nt}. When selecting two target masks, we achieve the best performance (\ding{183}). When increasing $n_t$ to 3, the performance slightly drops (\ding{184}). Thus, we set $n_t$ to 2.

\subsection*{S6.2 \; Impact of $n_d$ in DFG}

In VQ-SAM, $n_d$ determines the number of selected distractor masks for memory evolution. As shown in Tab.~\ref{tab:nd}, when using only one distractor for memory evolution, VQ-SAM shows best results (\ding{182}).

\subsection*{S6.3 \; Impact of $\gamma_k$ in Optimization}

\vspace{-0.35em}
\setlength{\intextsep}{0pt}
\setlength{\columnsep}{9pt}%
\begin{wraptable}{r}{0.450\textwidth}
\setlength{\tabcolsep}{9pt}
	\centering
	\renewcommand{\arraystretch}{1.0}
    \caption{Ablation study on $\gamma_1$ and $\gamma_2$.}
	\scalebox{0.65}{
    \begin{tabular}{ccccccc}
    \Xhline{1.2pt}
    & $\gamma_1$ & $\gamma_2$ & mSTIoU & mTIoU & Rec & Succ \\
   \hline\hline
   \ding{182} & 0.5 & 0.5 & 24.2 & 27.0 & 40.4 & 39.0 \\
   \rowcolor{cyan!10}
   \ding{183} & 0.5 & 1.0 & 26.0 & 29.6 & 43.6 & 42.1 \\
   \ding{184} & 1.0 & 0.5 & 24.5 & 27.4 & 41.2 & 39.9 \\
   \ding{185} & 1.0 & 1.0 & 25.3 & 28.7 & 42.2 & 40.8 \\
    \Xhline{1.2pt}
    \end{tabular}}
	\label{tab:loss_weight_ablation}
\end{wraptable}
In VQ-SAM, multiple stages are jointly optimized, and the loss of each stage is controlled by $\gamma_k$ (see Sec. S5). To study the impact of $\gamma_k$ (\ie, $\gamma_1$ and $\gamma_2$ for the two stages in VQ-SAM, we conduct an ablation study in Tab.~\ref{tab:loss_weight_ablation}. When setting $\gamma_1$ and $\gamma_2$ to 0.5 and 1.0, respectively, we achieve the best results (see \ding{183}).

\subsection*{S6.4 \; Visualization of Selected Target and Distractor Cues}

\setlength{\columnsep}{10pt}%
\setlength\intextsep{5pt}
\begin{wrapfigure}{r}{0.65\textwidth}
\centering
\includegraphics[width=0.65\textwidth]{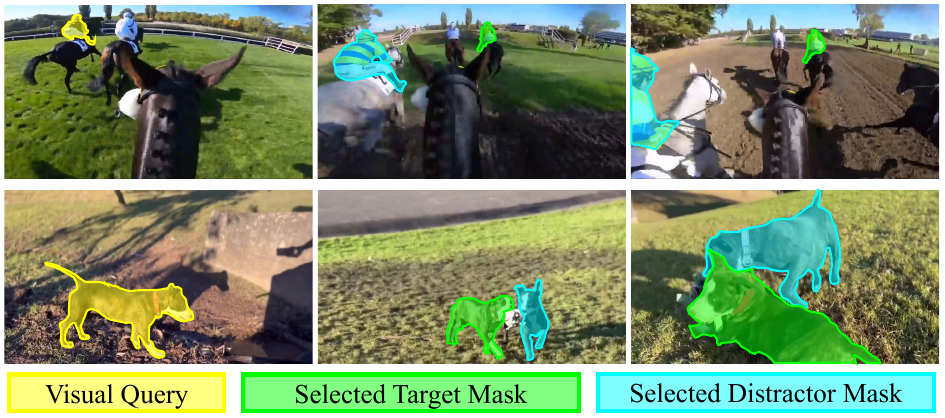}
\caption{Selected target and distractor cues.}\vspace{-5mm}
\label{fig:selected_cues}
\end{wrapfigure}
VQ-SAM aims to exploit target and distractor cues from the video for evolving the memory for better visual query segmentation. In Fig.~\ref{fig:selected_cues}, we show the selected target and distractor in the video. From Fig.~\ref{fig:selected_cues}, we can observe that, VQ-SAM can accurately identify target-specific and background distractor regions in the videos, which are used to enhance the discriminability of evolved memory.

\subsection*{S6.5 \; Comparison of Attention Maps}

\begin{figure}[!t]
   \centering
   \includegraphics[width=0.95\textwidth]{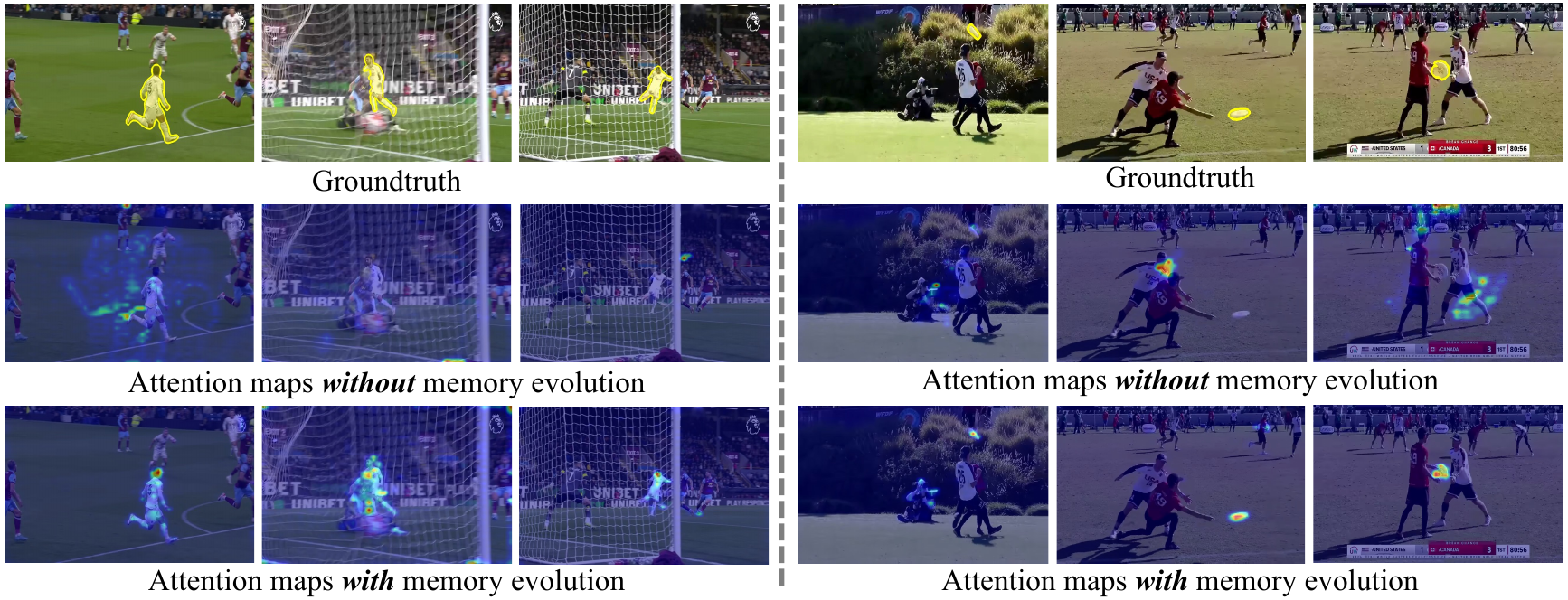}
   \caption{Comparison of attention maps with and without memory evolution. From this figure, we can clearly observe that the attention maps of with memory evolution can better focus on the target object for segmentation and localization.}
   \label{fig:att_map_comp}
\end{figure}

To analyze the efficacy of our memory evolution for visual query localization, we show the attention maps in the memory attention module with and without using our approach in Fig.~\ref{fig:att_map_comp}. From Fig.~\ref{fig:att_map_comp}, we can see that, our method with memory evolution using target and distractor cues is able to better focus on queried target in the video, leading to better localization and segmentation performance. 

\subsection*{S6.6 \; Qualitative Results}

\begin{figure}[!t]
   \centering
   \includegraphics[width=0.95\textwidth]{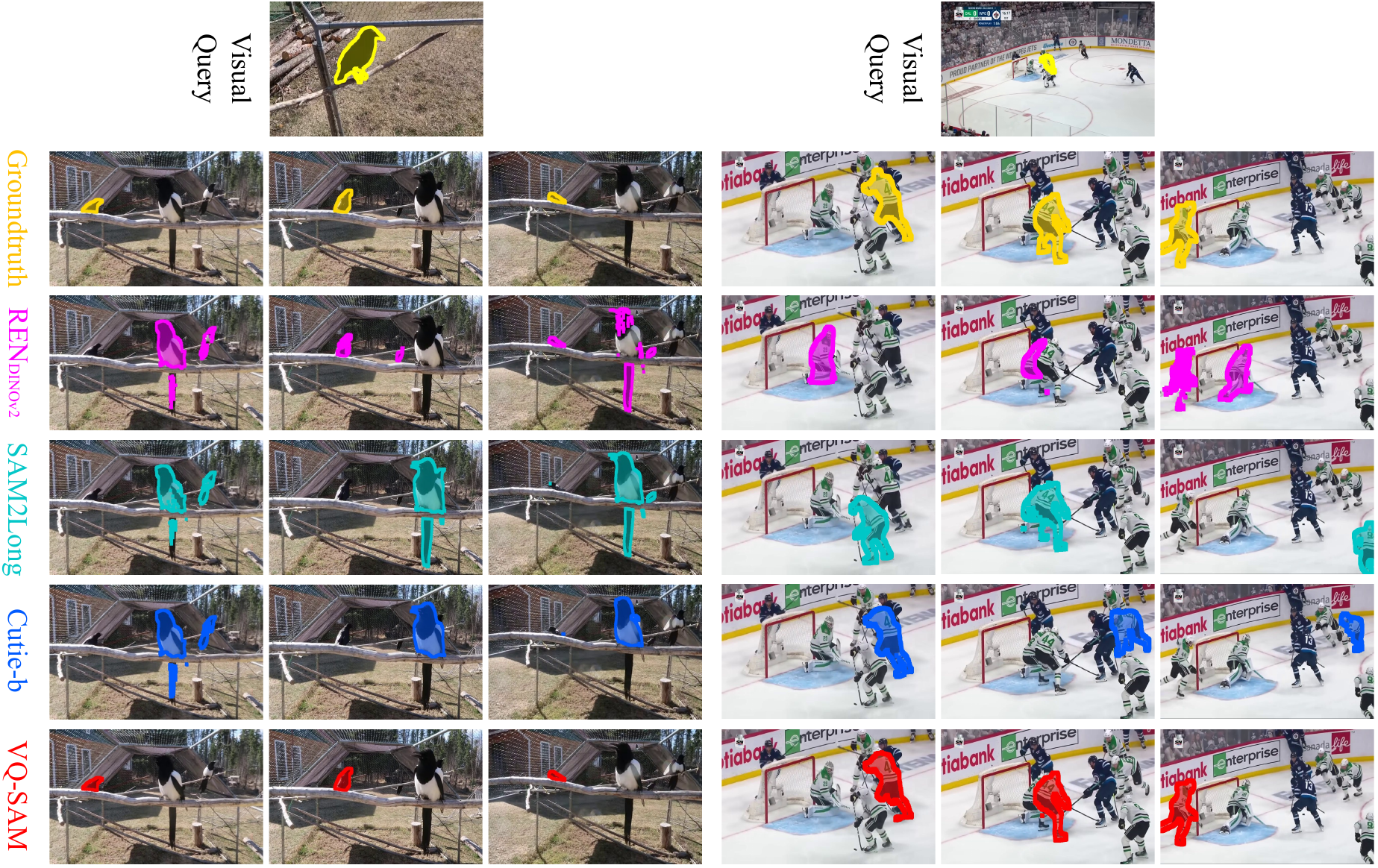}
   \caption{Qualitative results of our VQ-SAM and representative methods.}
   \label{fig:qual_comp}\vspace{-5mm}
\end{figure}

In order to provide further analysis of our VQ-SAM, we demonstrate the qualitative results of its segmentation and compare VQ-SAM against representative methods in Fig.~\ref{fig:qual_comp}. From the shown visualizations in Fig.~\ref{fig:qual_comp}, other methods struggle to locate and segment the target object accurately in the video. In contract, our VQ-SAM successfully segments the target across frames, showing the efficacy of memory evolution and spatial-temporal modeling in our approach.

\subsection*{S6.7 \; Annotation Accuracy Analysis} To analyze annotation accuracy, we randomly select 150 videos in VQS-4K and ask an independent group for re-labeling. After this, we compute the Intersection over Union (IoU) of the new and original annotations, and the IoU is 0.92, showing the reliability of our annotations.

\renewcommand{\arraystretch}{1.05}
\begin{table}[!t]
\setlength{\tabcolsep}{5.5pt}
  \centering
  \caption{Comparison on model efficacy and complexity. Please note, all methods are assessed on the same platform with 4 Nvidia A6000 GPUs (48G) for fair comparisons.}\vspace{2mm}
  \resizebox{0.92\textwidth}{!}{
    \begin{tabular}{rccccccc}
    \Xhline{1.2pt}
    \rowcolor[HTML]{f2f3f4}
    Methods & Resolution & FLOPs/frame & FPS & Inf. GPU Mem.  & mSTIoU & mTIoU \\
    \hline\hline

    Cutie-b\textcolor{gray}{\scriptsize{[CVPR'24]}}~\cite{cheng2024putting} 
    & 480$\times$854 & 185\,G & 36 & 3.0\,G & 18.5 & 21.0 \\

    OASIS\textcolor{gray}{\scriptsize{[ICCV'25]}}~\cite{qin2025structure} 
    & 480$\times$854 & 162\,G & 48 & 3.2\,G & 15.8 & 21.3 \\

    LOMM\textcolor{gray}{\scriptsize{[ICCV'25]}}~\cite{lee2025lomm} 
    & 448$\times$448 & 94\,G & 10 & 2.5\,G & 12.5 & 12.7 \\

    SAM 2\textcolor{gray}{\scriptsize{[ICLR'25]}}~\cite{RaviGHHR0KRRGMP25} 
    & 1024$\times$1024 & 844\,G & 22 & 5.1\,G & 12.9 & 13.0\\

    SAM2.1++\textcolor{gray}{\scriptsize{[CVPR'25]}}~\cite{videnovic2025distractor} 
    & 1024$\times$1024 & 903\,G & 16 & 5.5\,G & 12.5 & 14.6 \\

    SAM2Long\textcolor{gray}{\scriptsize{[ICCV'25]}}~\cite{ding2025sam2long} 
    & 1024$\times$1024 & 912\,G & 19 & 5.3\,G & 18.6 & 24.4 \\

    SAM 3\textcolor{gray}{\scriptsize{[ICLR'26]}}~\cite{carion2025sam} 
    & 1280$\times$1280 & 1385\,G & 15 & 10.0\,G & 13.3 & 14.1 \\

    VQLoC\textcolor{gray}{\scriptsize{[NeurIPS'23]}}~\cite{jiang2023single} 
    & 448$\times$448 & 228\,G & 36 & 4.0\,G & 10.1 & 21.5 \\

    PRVQL\textcolor{gray}{\scriptsize{[ICCV'25]}}~\cite{fan2025prvql} 
    & 448$\times$448 & 264\,G & 30 & 5.0\,G  & 13.6 & 22.1 \\

    REN\textcolor{gray}{\scriptsize{[NeurIPS'25]}}~\cite{khosla2025ren} 
    & 672$\times$672 & 511\,G & 12 & 4.0\,G & 16.8 & 23.6 \\
    
    \hline

    \rowcolor[HTML]{E7E7FF}
    \textbf{VQ-SAM} (ours) 
    & 768$\times$768 & 458\,G & 29 & 6.8\,G & 26.0 & 29.6 \\

    \Xhline{1.2pt}
    \end{tabular}}
  \label{tab:vqsam_efficiency}%
\end{table}

\subsection*{S6.8 \; Efficiency and Complexity Analysis}

To further analyze the efficiency and complexity of our method, we conduct a comparative analysis involving FLOPs, FPS, and inference GPU memory. As shown in Tab.~\ref{tab:vqsam_efficiency}, our method is more efficient than many existing methods such as SAM2, SAM2.1++, SAM2Long, SAM 3, and REN, and meanwhile achieves better performance with 26.0\% mSTIoU and 29.6\% mTIoU scores. These results demonstrate that VQ-SAM achieves a favorable balance between efficiency and accuracy.

\subsection*{S6.9 \; Analysis for Dataset Validation}

To enable in-depth understanding of our dataset, we further analyze it from the following aspects.

\textbf{Analysis on Category Diversity.} VQS-4K aims to provide a diverse platform for VQS. To this end, VQS-4K is organized into 19 coarse semantic groups comprising 222 fine-grained object categories, covering a broad of spectrum of object types, including human, animals, vehicles, electronics, household items, sport equipment, tools, and many others (please kindly see Fig.~\ref{fig:class-dist} for the category organization). The dataset also exhibits a long-tailed category distribution (we provide the category frequency distribution alone with the code and data), containing both common and rare categories. This covers varying levels of semantic difficulty while better capturing the diversity of objects encountered in real-world videos. As a result, VQS-4K enables a comprehensive evaluation of model robustness across a wide range of object appearances and semantic concepts.

\textbf{Analysis on Object Scale.} VQS-4K contains targets of different scales. As shown in Fig.~\ref{fig:stats} (d) in the paper, the object mask areas in VQS-4K exhibit a broad long-railed distribution, ranging from very small to large objects. In addition, based on the average object mask area of each video, we further categorize the dataset into Small (average mask in the range of (0,$3.6\times10^3$)), Medium (average mask in the range of [$3.6\times10^3$,$4.0\times10^4$]), and Large (average mask in the range of greater than $4.0\times10^4$) subsets using predefined area thresholds. Accordingly, across the entire VQS-4K dataset, 29.6\% (1,216), 51.1\% (2,117), and 18.9\% (778) of videos belong to Small, Medium, and Large, respectively. These statistics demonstrate that VQS-4K covers a broad spectrum of object scales rather than being biased toward a particular size, introducing varying levels of segmentation difficulty from challenging small objects with limited visual cues to larger objects with richer appearance information.

\textbf{Analysis on Video Length.} We display the distribution of video length of VQS-4K in Fig.~\ref{fig:stats} (a) in the paper. As shown, the video durations range from 5.3 seconds to over 80 seconds, with an average duration of 52.9 seconds. The duration distribution covers both short and long videos, resulting in substantially different temporal search spaces. While shorter videos typically require searching over fewer frames, longer videos introduce more background content, distractors, and temporal variations, making long-term visual query retrieval considerably more challenging. This introduces varying levels of temporal difficulty in VQS.

\textbf{Analysis on Sparse Object Occurrence.} VQS-4K contains diverse object occurrence patterns. As shown in Fig.~\ref{fig:stats} (c), the queried object may appear a varying number of times in the video, with 2.9 occurrences on average, indicating diverse occurrence frequencies across videos. However, occurrence count alone does not fully capture how sparsely the queried object is distributed throughout a video. We therefore compute the temporal occupancy ratio, defined as the proportion of frames in which the queried object is visible. A lower occupancy ratio indicates that the queried object appears in only a small portion of the video, corresponding to greater temporal sparsity and increased segmentation difficulty. Across VQS-4K, the average temporal occupancy ratio is 18.3\%, demonstrating that the object occurrence is overall sparse in VQS-4K and the queried objects are visible in only a limited portion of many videos. Together, the occurrence-count distribution (in Fig.~\ref{fig:stats} (c)) and temporal-occupancy analysis show that VQS-4K covers diverse levels of temporal occurrence sparsity, including queried objects that appear briefly, intermittently, or repeatedly throughout a video.

\textbf{Analysis on Video Source.} We source videos in VQS-4K from YouTube, because it is one of the largest video platform and provides diverse real-world videos covering a wide range of environments, viewpoints, activities, and recording conditions. During video collection, for each object category, we intentionally selected videos from diverse scenarios rather than similar videos from a single scene or context. This strategy helps reduce source/scenario bias and improves the diversity of the benchmark. We will clarify this data collection protocol in revision.

\subsection*{S6.10 \; Inference Details}

During inference, each full, untrimmed video is sequentially divided into non-overlapping clips (or temporal windows) of 7 frames, regardless of its duration. If the final clip contains fewer than 7 frames, it is padded by repeating its last frame. Each clip is then independently processed by VQ-SAM to generate frame-level predictions, which are concatenated in their original temporal order to reconstruct the predictions for the entire video. For the padded final clip, predictions corresponding to the repeated frames are discarded, ensuring that each original frame has exactly one prediction. Consequently, no temporal aggregation, duplicate masklet merging, or non-maximum suppression is required. Separated target occurrences are identified directly from the reconstructed frame-level predictions without additional post-processing. Since the reconstructed predictions preserve the original video length and temporal order, the padding and stitching procedure does not affect the computation of temporal evaluation metrics.

\renewcommand{\arraystretch}{1.05}
\begin{table}[!t]
  \centering
  \setlength{\tabcolsep}{5pt}
  \caption{Comparison under seen-category and unseen-category settings.}\vspace{1mm}
  \resizebox{0.92\textwidth}{!}{
    \begin{tabular}{rcccccccc}
    \Xhline{1.2pt}
          & \multicolumn{4}{c}{Seen-category}
          & \multicolumn{4}{c}{Unseen-category} \\
          \cmidrule(lr){2-5} \cmidrule(lr){6-9}
    Methods & mSTIoU & mTIoU & Rec & Succ
            & mSTIoU & mTIoU & Rec & Succ \\
    \hline\hline
    Cutie-b\textcolor{gray}{\scriptsize{[CVPR'24]}}~\cite{cheng2024putting}
        & 18.5 & 21.0 & 38.4 & 32.3
        & 16.6 & 18.5 & 33.6 & 28.3 \\
    OASIS\textcolor{gray}{\scriptsize{[ICCV'25]}}~\cite{qin2025structure}
        & 15.8 & 21.3 & 34.3 & 29.4
        & 14.7 & 19.6 & 31.6 & 28.7 \\
    LOMM\textcolor{gray}{\scriptsize{[ICCV'25]}}~\cite{lee2025lomm}
        & 12.5 & 12.7 & 26.1 & 25.8
        & 11.3 & 11.2 & 23.4 & 23.9 \\
    SAM 2\textcolor{gray}{\scriptsize{[ICLR'25]}}~\cite{RaviGHHR0KRRGMP25}
        & 12.9 & 13.0 & 26.7 & 26.3
        & 12.3 & 11.1 & 23.8 & 25.3 \\
    SAM2.1++\textcolor{gray}{\scriptsize{[CVPR'25]}}~\cite{videnovic2025distractor}
        & 12.5 & 14.6 & 28.8 & 25.8
        & 11.5 & 14.0 & 28.8 & 25.6 \\
    SAM2Long\textcolor{gray}{\scriptsize{[ICCV'25]}}~\cite{ding2025sam2long}
        & 18.6 & 24.4 & 38.6 & 32.5
        & 17.7 & 21.7 & 35.6 & 33.2 \\
    SAM 3\textcolor{gray}{\scriptsize{[ICLR'26]}}~\cite{carion2025sam}
        & 13.3 & 14.1 & 27.7 & 26.6
        & 10.4 & 12.0 & 22.8 & 20.4 \\
    VQLoC\textcolor{gray}{\scriptsize{[NeurIPS'23]}}~\cite{jiang2023single}
        & 10.1 & 21.5 & 31.7 & 23.2
        & 9.9 & 18.6 & 27.8 & 23.0 \\
    PRVQL\textcolor{gray}{\scriptsize{[ICCV'25]}}~\cite{fan2025prvql}
        & 13.6 & 22.1 & 32.6 & 26.9
        & 13.2 & 22.2 & 32.5 & 24.7 \\
    REN\textcolor{gray}{\scriptsize{[NeurIPS'25]}}~\cite{khosla2025ren}
        & 16.8 & 23.6 & 34.8 & 30.5
        & 14.6 & 21.0 & 30.5 & 26.9 \\
    \hline
    \rowcolor[HTML]{E7E7FF}
    VQ-SAM (ours)
        & 26.0 & 29.6 & 43.6 & 42.1
        & 25.0 & 28.4 & 41.9 & 40.5 \\
    \Xhline{1.2pt}
    \end{tabular}}
  \label{tab:unseen}%
  \vspace{-1em}
\end{table}%

\subsection*{S6.11 \; Category-Level Generalization}

To show the category-level generalization of our method, we conduct an additional unseen-category experiment, in which training and test sets do not have overlap in category. Specifically, we divide the 222 categories into 179 training categories, accounting for approximately 80\% of the categories and 3,282 videos, and 43 unseen test categories, accounting for approximately 20\% of the categories and 829 videos. Under this unseen-category setting, we retrain and evaluate all compared methods, including our VQ-SAM, using only the 179 training categories and testing exclusively on the 43 unseen categories. The results of the unseen-category experiment and comparison with the original category-overlapping (or seen-category) setting are summarized in Tab.~\ref{tab:unseen}. As shown in Tab.~\ref{tab:unseen}, all methods experience performance degradation under the more challenging unseen-category setting. Nevertheless, VQ-SAM consistently achieves the best performance across all evaluation metrics on unseen categories. By incorporating this unseen-category evaluation protocol, VQS-4K explicitly supports the evaluation of category-level generalization on unseen object categories, complementing the original evaluation setting.

\subsection*{S6.12 \; Difference between VQ-SAM and SAM 2}

To provide a strong baseline and practical guidance for future research, we develop VQ-SAM, a simple yet effective framework by extending SAM 2 explicitly for VQS. While VQ-SAM shares the similar idea of target and distractor memory modeling with prior work, our design is specifically tailored to the VQS task. Moreover, we introduce an adaptive memory generation module to dynamically integrate multiple memory cues for more effective memory representations. As demonstrated by the experimental results, VQ-SAM consistently outperforms existing methods on VQS-4K, establishing a strong baseline and providing useful guidance for future research.

\section*{S7 \; Ethical Statement and Maintenance of VQS-4K}

\textbf{Ethical Statement.} The construction of VQS-4K strictly follow ethical standards. All video sequences are collected under the \emph{Creative Commons} license, and are used for \emph{research purpose only}. However, we understand that the license of certain videos may change in the future. Once any notification of the change of video licenses is received, we will take appropriate actions to handle it.

\vspace{0.3em}
\noindent
\textbf{Maintenance.} VQS-4K will be hosted on the popular Github, which allows us to conveniently check feedback for improving VQS-4K via necessary maintenance. Besides, all experimental results of compared approaches and our VQ-SAM on VQS-4K will be released. Our ultimate goal is to offer a long-term and stable platform to foster research on visual query segmentation.

\section*{S8 \; Discussion on Limitation and Broader Impact}

\textbf{Limitation.}  Although our VQS-4K provides a large-scale and diverse benchmark for visual query segmentation, our work is limited in video length. While VQS-4K contains both short and relatively long videos, the average video length is still moderate compared with many real-world scenarios, such as long-duration surveillance and robotics. Extending VQS to substantially longer videos with more sparse, intermittent, and ambiguous target occurrences is an important future direction for VQS.

\vspace{0.3em}
\noindent
\textbf{Broader Impact.} This work introduces VQS, which aims to segment all pixel-level occurrences of a queried object in an untrimmed video. The proposed task, benchmark, and baseline may benefit a wide range of applications, including video retrieval, video editing, and robotics. By enabling more comprehensive and precise localization than box-based or last-occurrence localization, VQS can support downstream systems that require fine-grained video understanding.

At the same time, this capability also raises potential risks. In particular, the data and model could be misused for unauthorized tracking, surveillance, or monitoring of individuals and objects. Therefore, we emphasize that VQS-4K and VQ-SAM should be used only in ethical, privacy-preserving, and legally compliant settings. We hope that releasing this benchmark will encourage responsible research on VQS while promoting awareness of its potential societal risks.


\end{document}